\documentclass[msom,sglanonrev]{informs4}

\usepackage{eqndefns-left} 
\RequirePackage{tgtermes}
\RequirePackage{newtxtext}
\RequirePackage{newtxmath}
\RequirePackage{bm}
\RequirePackage{endnotes}

\OneAndAHalfSpacedXII 

\usepackage{algorithm}
\usepackage{algpseudocode}
\usepackage{tikz}

\usepackage{natbib}
 \bibpunct[, ]{(}{)}{,}{a}{}{,}%
 \def\bibfont{\small}%

\RequirePackage{amsmath}
\RequirePackage{amssymb}
\RequirePackage{bm} 
\RequirePackage{url}
\usepackage{multirow}
\usepackage{natbib}
\usepackage{graphicx}
\usepackage{subfigure}
\usepackage{makecell}
\usepackage{booktabs}
\usepackage{array}
\usepackage{xurl}
\usepackage{algorithm}
\usepackage{algpseudocode}
\algrenewcommand\alglinenumber[1]{\footnotesize\arabic{ALG@line}}
\usepackage{makecell}
\usepackage{diagbox}
\usepackage{appendix}

\usepackage{amsmath,amssymb,bbm}
\usepackage{natbib}
\usepackage{multirow}
\usepackage{subfigure}
\usepackage{makecell}
\usepackage{booktabs}
\usepackage{array}
\usepackage{tabularx}
\usepackage{tabulary}
\usepackage{caption}
\usepackage{booktabs}
\usepackage{url}
\usepackage{bm}
\usepackage{wrapfig}
\usepackage{lipsum}
\usepackage{dsfont}
\usepackage{color}
\usepackage{longtable}
\usepackage{relsize}
\usepackage{enumitem}

\usepackage{xr-hyper}
\usepackage[usenames,dvipsnames,svgnames,table]{xcolor}
\definecolor{myblue}{HTML}{0064A2}
\usepackage[colorlinks,linkcolor=blue,anchorcolor=blue,citecolor=blue,urlcolor=black]{hyperref}

\EquationsNumberedThrough    

\TheoremsNumberedThrough     
\ECRepeatTheorems  %

\MANUSCRIPTNO{}

\begin{document}
\RUNAUTHOR{He et al.}

\RUNTITLE{A SPORD Approach for Supply Chain Planning}

\TITLE{SPORD: A Simulation-Propose-then-OR-Dispose Approach for Supply Chain Planning}

\ARTICLEAUTHORS{%

\AUTHOR{Jiayin He}
\AFF{Department of Industrial Engineering, Tsinghua University, Beijing 100084, China,  \EMAIL{hejy23@mails.tsinghua.edu.cn}}

\AUTHOR{Yutong Pan, Sen Yang, Ningxuan Kang, Yongzhi Qi, Jianshen Zhang}
\AFF{JD.com, Beijing, China, 101111,  \EMAIL{panyutong5@jd.com,yangsen66@jd.com,kangningxuan@jd.com,qiyongzhil@jd.com,zhangjianshen@jd.com}}

\AUTHOR{Wei Qi}
\AFF{Department of Industrial Engineering, Tsinghua University, Beijing 100084, China,  \EMAIL{qiw@tsinghua.edu.cn}}

\AUTHOR{Zuo-Jun Max Shen}
\AFF{College of Engineering, UC Berkeley, Berkeley, CA 94720, USA; Faculty of Engineering, Faculty of Business and Economics, University of Hong Kong, Hong Kong, China,
\EMAIL{maxshen@berkeley.edu}}
} 

\ABSTRACT{%
For years, supply chain planning at e-commerce firms has operated as a collection of isolated projects. Each planning task from static network planning to dynamic warehouse assortment planning requires analysts to spend weeks building models from scratch, calibrating and persuading executives to act on outputs they cannot verify. Three barriers drive this: bespoke models proliferate because standardization is difficult (operational fragmentation); once unified, the combinatorial scale of millions of SKUs, thousands of nodes, and intricate routing logic exceeds what solvers can handle within a tight window (computational intractability); and a mathematically optimal solution still fails to be implemented if the executives do not trust it (implementation hurdle). To bridge this gap, we propose and implement the Simulation-Propose-then-OR-Dispose method, deployed as JD.com’s \textit{NetSim} platform. The central insight is decoupling: simulation proposes by generating and evaluating the full set of operationally valid candidate paths-absorbing all idiosyncratic business logic, while an integer program disposes by selecting the globally optimal subset. Computationally, matrix-vectorized CPU/GPU accelerated simulation achieves a 10-100 times speedup over serial methods, and a list scheduling algorithm reduces coupled-order processing from hours to minutes. Operationally, we establish a closed loop via an intelligent diagnosis engine. Since 2025, \textit{NetSim} has optimized end to-end services for over 20,000 suppliers, the cross-regional fulfillment rate dropped from 6.1\% to 4.9\%, and the average monthly carbon reduction is approximately 5,745 tCO2e. SPORD moves simulation from monitoring to active planning. The transparent outputs turn skeptical executives into engaged collaborators, and the modular architecture ensures that the next planning requires just configuration, not reconstruction.}

\KEYWORDS{Simulation, Supply chain management, Operations strategy, OM practice} 

\maketitle

\section{Introduction}
\label{introduction}
Modern e-commerce supply chains are characterized by massive scale, high dynamism, and intricate coupling, placing the cost-service trade-off at the center of strategic operations. In low-margin, high-velocity retail environments, stringent cost control is a profitability imperative \citep{zhou2025commerce}, while a high service level, primarily manifested as rapid and reliable delivery \citep{hu2024supercharged},  remains a cornerstone of customer loyalty. These objectives are often in inherent conflict, as pursuing lower costs can sometimes compromise the system's responsiveness and order fulfillment timeliness, and thus underscore the need for multi-echelon optimization to maximize overall efficacy while navigating the balance between cost and service.

\begin{figure}[H]
    \centering
    \includegraphics[width=0.8\linewidth]{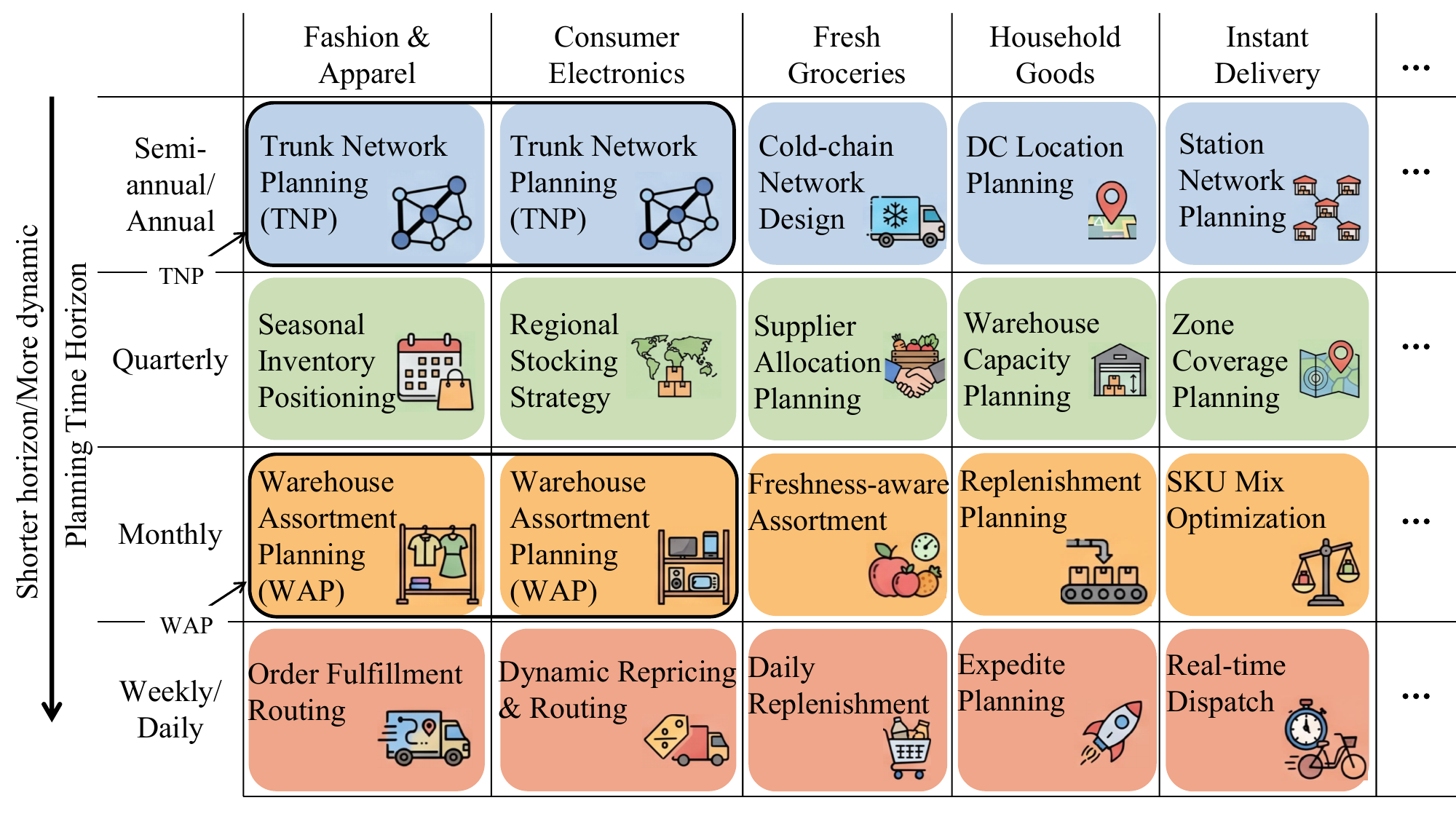}
    \caption{The Landscape of Supply Chain Planning (SCP) Problems at JD.com}
    \label{fig:business_units_time_horizon}
\end{figure}

Supply chain planning (SCP) is a pivotal framework for navigating this trade-off. Essentially, SCP encompasses a massive decision-making ecosystem that generally spans a spectrum from long-term static structural design (e.g., Trunk Network Planning (TNP), typically evaluated over a 6-to-12-month horizon) to short-term dynamic tactical planning (e.g., Warehouse Assortment Planning (WAP), adjusted on a weekly or monthly basis). To appreciate the scope of this heterogeneity, Figure \ref{fig:business_units_time_horizon} maps the SCP landscape along two dimensions: the horizontal axis enumerates representative business units within JD.com's ecosystem, each governed by distinct cost structures, service-level requirements, and product-handling rules, while the vertical axis stratifies planning time horizons from semi-annual structural decisions down to weekly or daily tactical adjustments. Each cell represents a distinct, concurrently running SCP problem. TNP and WAP, highlighted in Figure \ref{fig:business_units_time_horizon}, anchor almost opposite ends of this spectrum and serve as the two representative SCP problems examined in this paper.

\vspace{-0.5cm}
\vspace{-1cm}

\begin{figure}[H]
    \centering
    \includegraphics[width=0.8\linewidth]{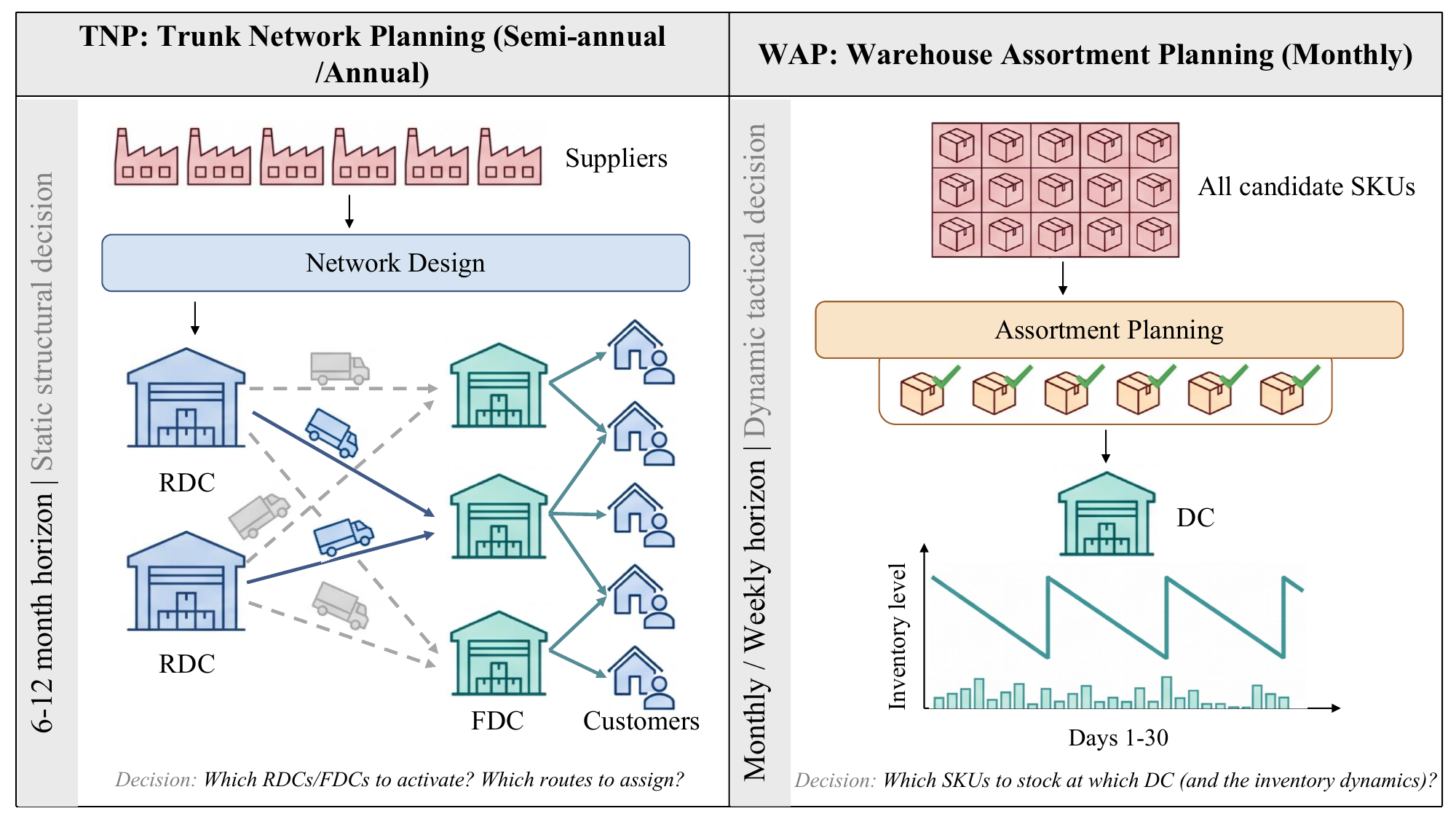}
    \caption{Illustration of TNP and WAP problem}
    \label{fig:TNP_WAP}
\end{figure}

For structural design (e.g. TNP), this planning process refers to (national or regional) multi-echelon end-to-end network planning for goods to distribute from suppliers (O) to customers (D). The planning output must be an optimal set of decisions on product flows, candidate warehouse site selection, network hierarchy \& coverage relationships, and transportation modes. For short-term tactical planning (e.g. WAP), this process determines the pairing relationship between SKUs and warehouses at each decision-making stage. Figure \ref{fig:TNP_WAP} zooms in on what TNP and WAP do in detail.

\vspace{-0.5cm}
\vspace{-1cm}

In practice, a massive e-commerce ecosystem like JD.com must continuously solve hundreds of such diverse planning variants. This complexity echoes the significant challenges encountered by global logistics leaders like Amazon \citep{shavarani2018application,jeong2022optimal} or DHL \citep{dang2024innovative}. Specifically, even translating theoretical optimization into actionable industrial practice is further impeded by several primary bottlenecks (which we will formally categorize and elaborate upon in Section \ref{challenges}): (i) Operational fragmentation: Developing, solving, and maintaining isolated, bespoke models for every unique business scenario requires prohibitive engineering overhead (detailed in Challenge I).
(ii) Computational intractability: Given millions of discrete decisions and tightly interacting operational nodes, traditional exact methods fail to compute a global optimum within operational time limits (detailed in Challenge I).
(iii) Implementation hurdles: A significant trust deficit stems from relying on opaque mathematical algorithms, hindering validation and executive buy-in (detailed in Challenge II).
(iv) Counterfactual deficit: The reliance on biased historical data fails to capture the nuances of unobserved scenarios, limiting the predictive power of traditional approaches (detailed in Challenge III). Together, these multifaceted challenges create a substantial barrier to realizing the full potential of SCP. 

A promising direction is to transition toward large-scale parallel simulation, which naturally absorbs complex business rules without bespoke mathematical reformulations and leverages massive parallel computing for rapid evaluation. Seamlessly integrating simulation with prescriptive optimization, however, remains the critical architectural challenge that motivates this paper.

In response to the aforementioned challenges, we propose a Simulation-Propose-then-OR-Dispose (SPORD) method. SPORD is a general-purpose methodology designed to establish a unified framework across all scales (e.g., from TNP to WAP) of supply chain planning. At its core, 
SPORD achieves this goal by standardizing both static physical routes (i.e. end-to-end logistics) in problems like TNP and dynamic inventory trajectories in problems like WAP into a universal candidate path representation. This design handles complex business logic by decoupling feasibility evaluation and cost/service evaluation (simulation) from prescriptive selection (optimization). For dynamic WAP specifically, a ``candidate path" represents a simulated temporal trajectory of inventory flow for a given SKU-to-warehouse (for specific customer demand) assignment over a specified horizon $T$. Tracing this inventory trajectory via simulation serves as an evaluation mechanism for assortment decisions: it dynamically measures the operational costs and service impacts of each candidate assignment under realistic inventory constraints, capturing replenishment dynamics, capacity limits, and complex mismatch costs that simplistic analytical assumptions cannot accommodate. A question at this point is whether exhaustively evaluating candidate paths is computationally feasible, given that the space of SKU-warehouse-trajectory combinations is in principle exponential. SPORD addresses this through a two-stage tractability design. First, the Initialization Layer applies expert-defined logical masks and flow discretization to prune the raw combination space, retaining only those paths that satisfy local operational validity conditions (e.g., permissible storage rules, minimum order quantities, routing feasibility). This transforms an intractable combinatorial explosion into a finite, business-meaningful candidate set. Second, a massively parallel simulation engine which leverages distributed computing and GPU-accelerated matrix operations evaluates the candidate paths simultaneously within operational time windows. In this way, both spatial and temporal paths encapsulate their corresponding operational costs and service metrics under different evaluation strategies. Subsequently, a simple, transparent Operations Research (OR) model disposes of the complexity by selecting the globally optimal subset of paths from this simulated candidate space. By doing so, SPORD eliminates the need for start-from-scratch mathematical formulations across different business domains.

The contributions of this paper are threefold: 
\begin{enumerate}
    \item This work represents a practice-based attempt to address the persistent, structural challenges of industrial-scale supply chain planning, i.e., operational fragmentation, computational intractability, and a deep trust deficit between algorithmic outputs and business stakeholders. The proposed SPORD method offers an architectural response: by standardizing both static network fulfillment routes and dynamic inventory trajectories into a universal candidate path representation, SPORD establishes a unified planning framework that spans the full spectrum from long-horizon trunk network design to short-cycle warehouse assortment decisions. This design decouples the complexity of feasibility evaluation from prescriptive optimization, eliminating the need for start-from-scratch mathematical formulations across diverse business units. The result is a system with generalizability, maintainability, and transparency.
    \item Technically, we pioneer a parallel simulation architecture based on Matrix Computing. By vectorizing complex discrete events, we transform the simulation process into matrix operations handled by CPU/GPU stacks, achieving a computational acceleration of 10 to 100 times. The inherent efficiency of parallel simulation extends beyond specialized hardware. Apache Spark generates high-quality solutions rapidly, even without GPU dependency. Furthermore, to overcome the computational bottlenecks incurred by coupled orders with precedence constraints, we integrate a list-scheduling algorithm based on topological sorting.
    \item We demonstrate the operational viability of the SPORD method through its industrial implementation as \textit{NetSim}, JD.com's enterprise supply chain planning platform. \textit{NetSim} simulates JD.com's entire supply chain at order-level resolution, digitally replicating the complete spectrum of distribution networks, forecasting, procurement, replenishment, transshipment, and fulfillment. Deployed to optimize end-to-end services for more than $20,000$ suppliers since 2025, the platform provides a reproducible solution for enterprise-level supply chain digital transformation. Empirical results confirm that SPORD effectively navigates the cost-service trade-off at national scale: the cross-regional fulfillment rate dropped from 6.1\% to 4.9\% (a 20\% reduction), boosting the consumer experience. WAP applications have delivered over 73 million dollars in annual cost savings, and the average monthly carbon reduction stands at approximately 5,745 tCO2e—comparable to preventing a 9.6-meter vehicle from traveling over 15 million kilometers. Granular evidence across ten business units over eighteen months further demonstrates that savings compound as platform coverage expands.
\end{enumerate}

The remainder of this paper is organized as follows. Section \ref{SCP_discuss} reviews the evolution of SCP, identifying the critical industrial bottlenecks that motivate our SPORD framework. Section \ref{NetSim_paraphase} formalizes the SPORD method, detailing its core pipeline from the simulation-based path generation to the final optimization. To show the system's tractability for hyper-scale networks, Section \ref{sec:scalability} elaborates on our computational breakthroughs, specifically parallel simulation and list scheduling algorithm. Section \ref{cases} demonstrates the generalizability of our approach through two deployments at JD.com, while illustrating how visualization and intelligent diagnosis effectively close the operational loop. Finally, Section \ref{analysis} summarizes managerial insights and concludes the paper.

\section{Retrospect and Rethinking Supply Chain Planning}
\label{SCP_discuss}
\vspace{0.2cm}
\subsection{Evolution of Supply Chain Planning}
\label{evolution_of_SCP}

\textbf{Phase I: Arc-Based Multi-Commodity Models}. In its classical form, fulfillment planning is modeled as a Multi-Commodity Network Flow problem. The decision variables are defined on physical arcs (e.g., $x_{ijk}$ denoting flow of commodity $k$ on link $(i,j)$), and the objective is to optimize flow allocation subject to capacity and conservation constraints. Foundational works, such as \cite{iri1969network}, establish the algorithmic basis for these flows, demonstrating their theoretical efficiency for large-scale transport scheduling and personnel assignment. To address the complexities of multi-echelon systems, \cite{van2005integrated} extend this into a monolithic framework. They formulate a polynomial-time algorithm that integrates production capacities, inventory lot-sizing, and transportation arcs into a single dynamic program. This model treats the supply chain as a synchronized rigid pipeline. As supply chains face increasing uncertainty, the rigidity of deterministic arcs becomes apparent. \cite{klibi2010design} have critically reviewed this era, arguing that standard arc models fail to achieve value-creating robust networks because they cannot easily encode stochastic disruptions or rule-based constraints without fundamentally altering the model structure. In response, \cite{bertsimas2013robust} introduce Robust and Adaptive Network Flows. This method immunizes arc flows against node/link failures via max-flow/min-cut duality, though this adds significant computational complexity. Most recently, \cite{ramos2022arc} present an exact arc flow model based on a graph to solve the multitrip production, inventory, distribution, and routing problem with time windows and validates the effectiveness of the proposed approach through a series of experimental tests. However, this reliance on arc-based constraints tightly couples business logic with mathematical formulations. Consequently, routine operational updates require analysts to manually recode the core model structure, causing significant deployment delays.

\textbf{Phase II: Path-Based Service Network Design}. To escape the granular rigidity of link-by-link decision making, researchers shift to path-based formulations. In this paradigm, the fundamental decision unit is no longer a physical arc, but a complete \textit{fulfillment chain} (e.g., Supplier$\rightarrow$Cross-dock$\rightarrow$Customer). \cite{crainic2000service} reviews this approach as Service Network Design, emphasizing that tactical planning is fundamentally about selecting the optimal configuration of services rather than merely routing volume through a fixed grid. This structural shift allows models to incorporate complex, non-additive path attributes that arc-based models could not easily handle. For instance, \cite{verter2008path} utilize path formulations to manage hazardous material risks, where the feasibility of a route is determined by a holistic measure of population exposure rather than a simple sum of link costs. Similarly, \cite{heydari2017path} apply path-based flows to railcar distribution, successfully capturing intricate blocking rules and train capacity constraints that previous models struggled to handle. To solve these exponentially large path sets, algorithms like Column Generation become the standard. The efficiency of this decomposition depends on solving the pricing problem to construct new valid paths. When business rules involve highly customized or interdependent logic, this sub-problem becomes analytically intractable. A recent study by \cite{guan2024path} illustrates this challenge: when modeling multi-modal transit systems with adoption awareness (where path validity depends on complex user behavior logic), standard mathematical generation fails, forcing researchers into computationally heavy enumeration or complex reformulations. This suggests that as business logic becomes more arbitrary and rule-based, purely mathematical methods struggle to generate feasible paths efficiently. 

\textbf{Phase III: The Rise of the Simulation-Optimization Era}. To address the business logic that purely mathematical problems struggled to handle, the field turned to Simulation-Optimization. This paradigm generally follows two distinct methodological streams. The first stream utilizes simulation to estimate uncertainty or approximate gradients through discrete event simulation, system dynamics simulation, etc., and then guides an OR model or an optimization algorithm. Recent advancements focus on improving the efficiency of this coupling by enhancing the ability to simulate complex dynamics. For instance, \cite{hogdahl2023combined} employ simulation to predict delay distributions, which are then fed into an optimization model for robust railway timetabling. Similarly, \cite{xu2023gradient} propose a multi-resolution approximation method, using simulation to construct stochastic gradients for optimizing complex systems (perform gradient search in the decision space). In the realm of solver architecture, \cite{cooper2020pymoso} introduce PyMOSO to solve problems on integer lattices using pseudo-gradients, while \cite{chang2025designing} have developed the ParMOO framework to handle multi-objective problems via parallel simulation execution. The second stream follows a guess-and-check logic. Here, the optimizer (e.g. (D)RL agent) proposes a candidate solution, which is then run through the simulation to evaluate its quality, followed by adjustments based on statistical feedback. With the rise of Industry 4.0, this concept also develops into the Digital Supply Chain Twin (DSCT) which has the potential to evaluate a given solution precisely. \cite{ivanov2021digital} theorize the DSCT as a real-time, computerized model capable of mapping physical network states to managing disruption risks. \cite{ivanov2024conceptualisation} further formalizes a 7-element framework for DSCTs, arguing that modern twins must integrate human-AI interactions to handle complex socio-technical tasks. 

\subsection{Challenges}
\label{challenges}
While the evolution demonstrates significant methodology progress, substantial gaps remain when translating these paradigms into industrial practice. The intrinsic limitations of current simulation engines, specifically regarding computational scalability and prescriptive power, are compounded by the complexity of human-algorithm interaction. This section details these multifaceted challenges.

\vspace{-0.2cm}
\subsubsection{Challenge I: The Dilemma between generalizability and customization.}

In practice, the existence of highly tailored strategies and constraints across business departments, such as varying service-level requirements for different products or distinct cost structures in different regions, has traditionally necessitated the development of bespoke network plans. Each business unit imposes distinct constraints—fashion categories prioritize inventory turnover, instant-delivery stations enforce brand quotas and order-value thresholds, and irregular items require bin-packing analyses. This proliferation of bespoke requirements, while locally precise, produces prohibitive engineering overhead and low delivery efficiency. A network planning project often requires a dedicated team of analysts to spend several weeks to months on data preparation, model calibration, and partner communication. This prolonged timeline prevents rapid responses to market and business changes. More critically, the business logic does not guarantee desirable mathematical properties such as convexity or Lipschitz continuity, which are often prerequisites for existing exact or heuristic algorithms. Therefore, many classical optimization algorithms are not directly applicable or may fail to find a good solution within a reasonable time \citep{shen2011reliable,seyhan2018new}. Consequently, this inefficiency incurs a critical need for a more scalable, automated solution. Transitioning to a universal, automated framework offers a pathway to scalability but also introduces a severe computational trade-off. For a single TNP instance, candidate path combinations across warehouses, SKUs, and routing rules can reach tens of millions, making the selection problem a large-scale combinatorial challenge.

\vspace{-0.2cm}
\subsubsection{Challenge II: The Gap Between Recommended Solution and Practical Implementation.}

Even after deriving a mathematically optimal solution, a significant chasm often remains before practical implementation can be realized. This gap is primarily driven by the importance of establishing trust and enabling validation within a compressed decision window. Business leaders rarely accept an opaque algorithmic solution outright. They require the ability to compare the proposed solution against a set of neighboring alternatives (e.g., by swapping one warehouse selected) across a range of familiar supply chain KPIs to understand its robustness and justify the decision. This requires not only accuracy but also rapid execution. Therefore, despite the roughly six-month life-cycle of a network plan, the decision-making process itself must be compressed to almost hour level to support rapid, interactive what-if analysis and foster buy-in by business departments. This requirement is amplified by the rise of generative AI: Large Language Models (LLMs) and autonomous agents that lack physical grounding routinely produce plausible but operationally infeasible outputs. A transparent simulation engine encoding supply chain physics and business rules is therefore essential as a verifiable ground truth for any algorithmic recommendation.

\vspace{-0.2cm}
\subsubsection{Challenge III: The Inherent Bias of Historical Data.}
\vspace{-0.1cm}

The third hurdle stems from the intrinsic limitations of historical operational data, which restricts the applicability of conventional statistical analysis or purely data-driven machine learning approaches. Supply chain data is observational and suffers from severe selection bias. Historically, operational decisions were governed by legacy business rules. For instance, specific SKUs might have been exclusively assigned to a single regional distribution center, while alternative routing options were never exercised. Therefore, the resulting dataset is structurally censored: it records performance metrics (e.g., cost, lead time) only for these pre-existing, manually selected assignments. This creates a highly sparse observation matrix and the performance of unattempted product-warehouse combinations remains unknown. This phenomenon incurs a critical deficit in counterfactual reasoning: the analytical capability to determine ``what would have happened" regarding cost and efficiency had a product flowed through a different, untried path. Without physical grounding, supervised/imitation learning algorithms are difficult to extrapolate to out-of-distribution configurations, creating a self-reinforcing loop that replicates past decisions rather than discovering superior structures.

\vspace{-0.2cm}
\subsection{Rethinking Supply Chain Planning}
\label{rethinking}
\vspace{-0.1cm}

The challenges outlined above collectively point to a structural inadequacy in how supply chain planning has traditionally been conceived. In fact, the fragmentation of bespoke models, the computational intractability of monolithic formulations, the bias of historical data, and the persistent trust deficit between algorithmic outputs and business stakeholders are not isolated symptoms. They share a common root: the attempt to encode idiosyncratic, non-linear operational reality directly into mathematical structure.

Motivated by the inherent limitations of existing approaches, we, as scholars and practitioners in supply chain management, posit two fundamental questions:

$\bullet$
\textit{In an era of burgeoning computational power, what fundamental properties must a planning methodology possess for next-generation supply chain planning?}

Based on our theoretical understanding and extensive field experience, we identify an urgent need for a next-generation, ultra-resolution simulation tool combined with OR models, the design of which must be strictly mapped to overcoming specific industrial hurdles. First, to resolve the conflict between operational fragmentation and large-scale application (Challenge I), the method must demonstrate \textit{generalizability \& maintainability}, enabling adaptation across diverse business scenarios via parameter configurations rather than hard-coding. Second, to break through the computational bottlenecks of large-scale combinatorial problems (Challenge I), it must possess \textit{scalability} to effectively handle millions of SKUs and complex topologies, alongside \textit{superior performance} to deliver high-quality solutions within hour-level decision windows. Third, to counter the descriptive but non-prescriptive nature of simulation and the bias of historical data (Challenge III), the system requires \textit{logical fidelity}, ensuring credibility by accurately capturing core business logic and enabling counterfactual reasoning rather than merely mimicking historical dataset. Finally, to mitigate the distrust in unexplainable solutions (Challenge II), \textit{usability} is essential to lower the understanding barrier for business experts of different backgrounds and foster buy-in. 

Simulation emerges as the natural embodiment of these properties. By digitally replicating the supply process, simulation enables low-risk \textit{what-if analysis} of various network configurations without disrupting actual operations. Furthermore, compared to currently popular reinforcement learning and deep learning technologies, it offers greater interpretability and business reliability, especially in decisions that involve hundreds of millions of dollars in economic benefits. The question then becomes architectural:

$\bullet$\textit{Which architectural design best embodies these principles?}

\begin{figure}[H]
    \centering
    \includegraphics[width=.8\linewidth,trim=1cm 1.5cm 1cm 2cm,clip]{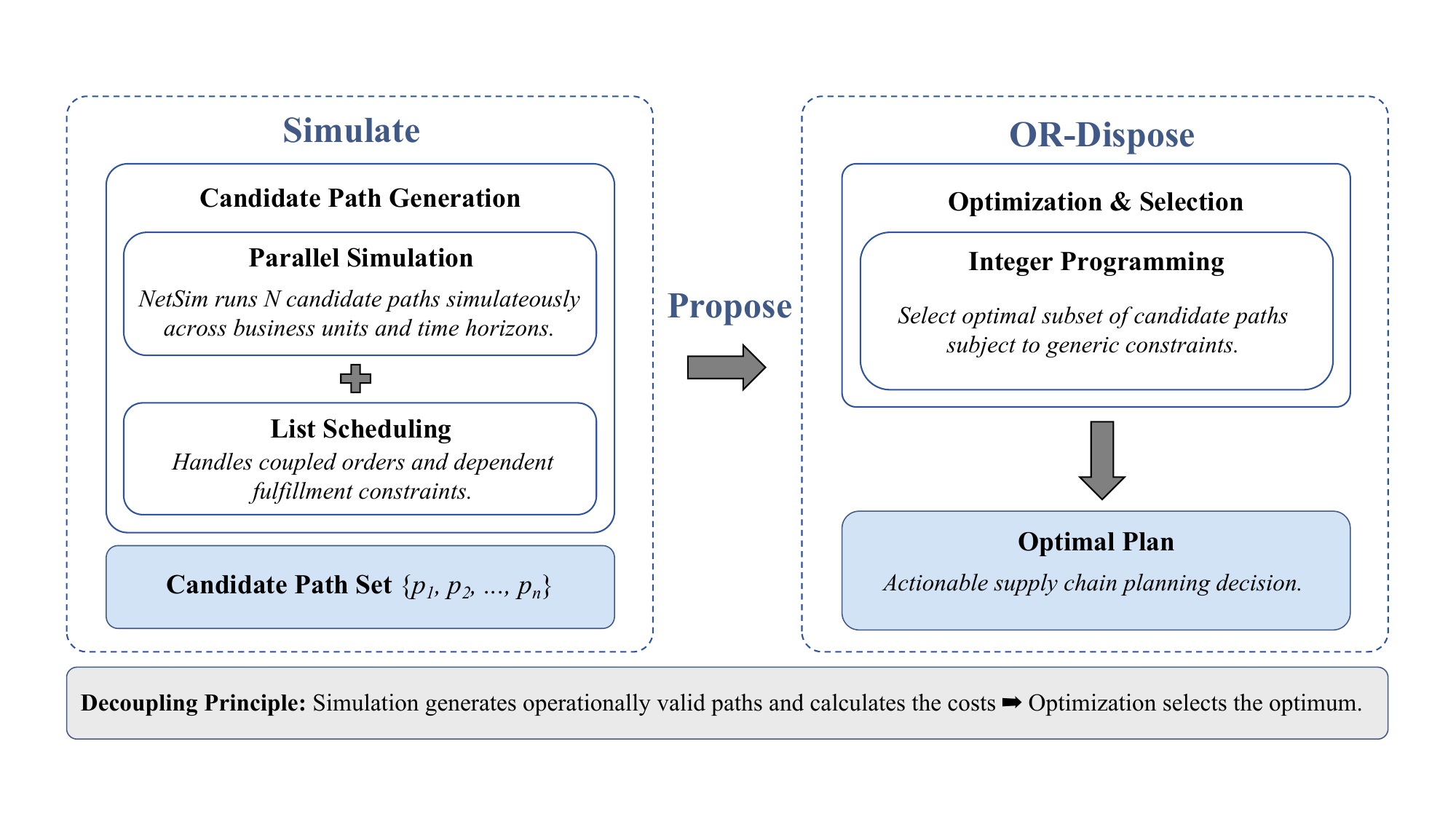}
    \caption{The SPORD Framework}
    \label{fig:S-then-O}
\end{figure}

We propose the Simulation-Propose-then-OR-Dispose (SPORD) paradigm as the answer. The central insight of SPORD is a deliberate decoupling: rather than asking an optimizer to simultaneously construct feasible solutions and respect complex operational constraints (a task that grows intractable at industrial scale), SPORD assigns these two responsibilities to separate, specialized components. Simulation is tasked with proposing: generating and evaluating the full set of operationally valid candidate paths, thus each carrying computed cost and service attributes. Optimization is tasked with disposing: selecting the globally optimal subset from this candidate space via a clean, transparent binary integer program. Because all idiosyncratic business logic has been absorbed upstream by simulation, the optimization problem is reduced to a structurally simple, universally solvable selection task—regardless of the complexity of the underlying operational domain (as shown in Figure \ref{fig:S-then-O}). This architecture is realized through a modular five-layer pipeline, detailed in Section \ref{NetSim_paraphase}.

\vspace{-.5cm}
\vspace{-1.1cm}

Two design properties are critical to making this paradigm work in practice: \textit{fidelity} and \textit{scalability}. The first concerns \textit{fidelity}: a simulation-based planning system is only as trustworthy as the candidate space it evaluates, which demands comprehensive coverage rather than sampling. The raw combinatorial space of candidate paths is exponentially large, but SPORD renders it tractable through a deliberate two-stage reduction. Expert-defined logical masks and flow discretization first prune the space to a finite, business-valid candidate set by encoding permissible routing relationships, order quantity bounds, and capacity thresholds. A massively parallel simulation engine then evaluates this pruned set simultaneously—leveraging distributed in-memory computation for independent paths, GPU-accelerated tensor operations for inventory-intensive temporal paths, and a list scheduling algorithm that eliminates thread starvation for precedence-coupled orders by dispatching jobs in critical-path priority order. Together, these mechanisms make exhaustive candidate evaluation not a theoretical ideal but a routine operational reality at JD.com's scale.

The second commitment concerns how uncertainty is handled to guarantee \textit{scalability}. Supply chain planning problems, particularly assortment and inventory problems, are frequently framed in the literature under stochastic demand assumptions, motivating formulations such as stochastic programming, sample average approximation, or robust optimization. We clarify that the planning problems addressed in this paper are treated as deterministic, and this is a deliberate design choice rather than a simplifying omission. In JD.com's operational setting, both TNP and WAP are solved against forecasted demand inputs that are fixed at the time of planning. Downstream business units generate demand forecasts over the planning horizon and submit them as committed inputs to the planning system. The planner's task is to identify the best structural or assortment decision given this input snapshot, not to hedge against an unrealized distribution of future demand. Several considerations justify this deterministic framing: Computationally, stochastic formulations such as SAA require solving across large scenario sets, which is prohibitive within the hour-level decision windows of industrial planning; robust optimization faces analogous scalability barriers and yields overly conservative solutions misaligned with margin-sensitive e-commerce operations. Structurally, the deterministic problem is itself far from trivial: its difficulty stems from heterogeneous non-convex business constraints and combinatorial scale, making stochastic approximation and iterative simulation-optimization loops both redundant and architecturally mismatched. Operationally, the deterministic setting yields transparent, auditable solutions that support systematic what-if analysis and align with the decision-making cadence of industrial workflows.

\section{The SPORD method}
\label{NetSim_paraphase}

To address the critical need for a high-fidelity, scalable planning tool, this section formalizes the SPORD method. To concretely illustrate the mechanics of this method, we will primarily utilize static TNP as our running example throughout this section. The application of SPORD to dynamic problems, specifically WAP, will be detailed later in the Industrial Applications (Section \ref{IA_application}) and Online Appendix \ref{appendix:applications}. We first present a comprehensive and representative multi-echelon network planning model (Section \ref{grand_model}) that exemplifies the computational and interpretability challenges faced by conventional monolithic approaches. To overcome the problem, we introduce the SPORD method and its modular components, detailing the sequential flow from generation to optimization.

\subsection{The Conventional Monolithic Model}
\label{grand_model}
To strictly describe the multi-echelon network planning problem, we start with a Mixed-Integer Linear Programming (MILP) model. 

Consider a multi-echelon supply chain network represented by a directed graph $G=(\mathcal{N},A)$, where $\mathcal{N}$ denotes the set of all nodes and $A$ represents the set of potential arcs. The node set $\mathcal{N}$ is partitioned into a set of candidate warehouse nodes $\mathcal{K}$ and a set of demand sink nodes $\mathcal{M}$, such that $\mathcal{N}=\mathcal{K}\bigcup\mathcal{M}$. The warehouse network is further structured hierarchically into disjoint layers indexed by $l\in\mathcal{L}=\{1,\dots,L\}$, where $\mathcal{K}_l\subseteq \mathcal{K}$ represents the subset of warehouses belonging to layer $l$. A subset of these warehouses, $\mathcal{F}\subseteq \mathcal{K}$, are designated as mandatory and must remain operational. The system manages the distribution of a set of heterogeneous commodities (SKUs), indexed by $s\in\mathcal{S}$. Each demand node $j\in \mathcal{M}$ has a specific demand $d_j^s$ for commodity $s$. The flow of goods through the network incurs two primary types of costs: (i) a fixed unit transportation and fulfillment cost $c_{ij}^s$, and (ii) a variable operational throughput cost $r_k$ incurred at warehouse $k\in\mathcal{K}$ for processing incoming flows. The feasibility of a link for a specific $s$ is a binary parameter $a_{ij}^s$. 

The planning problem involves two coupled decisions: network design and tactical flow allocation. First, we must determine the binary network configuration variables $y_k^l$, which equals to 1 if warehouse $k$ in layer $l$ is opened, and 0 otherwise. This selection is constrained by a predetermined number of facilities $h_l$ allowed for each layer. Second, we must determine the continuous flow variables $x_{ij}^s$, representing the quantity of commodity $s$ transported from node $i$ to node $j$.

The objective is to minimize the total cost, which consists of transportation costs along the arcs and variable operational costs at the warehouse nodes. This optimization is subject to flow conservation constraints at intermediate nodes, demand satisfaction at sink nodes, and capacity logic linking flows to open warehouses. Furthermore, the network must adhere to Service Level Agreements. We define a set of time thresholds $\mathcal{T}$, where for each threshold $\tau\in \mathcal{T}$, a minimum proportion $\rho_{\tau}$ of total demand must be satisfied within the transportation time $t_{ij}^s$ specified by the set $\mathcal{A}_{\tau}$. $\mathcal{A}_\tau = \{ (i,j,s) \mid t_{ij}^s \le \tau \}$ represents the set of arcs satisfying the time threshold $\tau$. Mathematically, this planning problem is specified as follows:
\vspace{-.85cm}
\begin{subequations}
\setlength{\jot}{0pt}
\begin{align}
    \min \quad &Z = \sum_{s \in \mathcal{S}} \sum_{i \in \mathcal{N}} \sum_{j \in \mathcal{N}} c_{ij}^s x_{ij}^s + \sum_{k \in \mathcal{K}} \left( r_k \sum_{s \in \mathcal{S}} \sum_{j \in \mathcal{N}} x_{kj}^s \right) \label{eq:objective}\\
    &\sum_{i \in \mathcal{N}} x_{ij}^s - \sum_{e \in \mathcal{N}} x_{je}^s = 0, \quad \forall j \in \mathcal{K} \setminus \mathcal{K}_1, \forall s \in \mathcal{S} \label{eq:flow_balance} \\
    &\sum_{i \in \mathcal{N}} x_{ij}^s = d_{j}^s, \quad \forall j \in \mathcal{M}, \forall s \in \mathcal{S} \label{eq:demand}\\
    &x_{ij}^s  \le M \cdot a_{ij}^s, \quad \forall i,j \in \mathcal{N}, \forall s \in \mathcal{S} \label{eq:valid_arc}\\
    &\sum_{j \in \mathcal{N}} \sum_{s \in \mathcal{S}} x_{kj}^s  \le M \cdot y_k^l, \quad \forall k \in \mathcal{K}_l,\ l \in \mathcal{L} \label{eq:bigM_out} \\
    &\sum_{i \in \mathcal{N}} \sum_{s \in \mathcal{S}} x_{ik}^s  \le M \cdot y_k^l, \quad \forall k \in \mathcal{K},\ l \in \mathcal{L} \label{eq:bigM_in}\\
    &\sum_{k \in \mathcal{K}_l} y_k^l = h_l, \quad \forall l \in \mathcal{L} \label{eq:layer_limit}\\
    &y_k^l = 1, \quad \forall (k,l) \in \mathcal{F}\cap\mathcal{K}_l \label{eq:mandatory}\\
    &\sum_{(i,j,s) \in \mathcal{A}_\tau} x_{ij}^s \ge \rho_\tau \cdot \sum_{s \in \mathcal{S}} \sum_{j \in \mathcal{M}} d_{j}^s, \quad \forall \tau \in \mathcal{T} \label{eq:sla}\\
    &x_{ij}^s \ge 0, \quad y_k^l \in \{0, 1\} \label{eq:def}\\
    & \bm{Ax} \le \bm{b}_{\text{custom}}(\mathcal{S}). 
    \label{eq:custom_constraints}
\end{align}
\end{subequations}

\vspace{-1cm}
Constraint \eqref{eq:flow_balance} ensures that for intermediate warehouses (excluding the first layer sources), the total inbound flow equals total outbound flow. Constraint \eqref{eq:demand} ensures demand satisfaction at sink nodes. Constraint \eqref{eq:valid_arc} restricts flow to valid arcs defined by the feasibility parameter ($a_{ij}^s$). Constraints \eqref{eq:bigM_out} and \eqref{eq:bigM_in} are Big-M constraints linking the continuous flow variables to the binary facility decisions; no flow is allowed through a closed warehouse. Constraint \eqref{eq:layer_limit} enforces the specific number of active warehouses required for each echelon layer. Constraint \eqref{eq:mandatory} ensures mandatory warehouses are selected. Constraint \eqref{eq:sla} mandates that the proportion of total network flow executed within the time limit $\tau$ must meet the service level target $\rho_\tau$. Constraint \eqref{eq:def} is consistent with the variable definitions. Constraint \eqref{eq:custom_constraints} represents a massive set of heterogeneous business rules which are defined by specific SKU-related department.

The monolithic model embeds numerous operational details that directly impact network topology and its performance. While theoretically sound, directly solving the above model in an industrial setting incurs several problems: (1) The Curse of Dimensionality: The variable space scales with $|\mathcal{S}|\times|\mathcal{K}|\times|M|$. The continuous nature of $x_{i,j}^s$ implies that any $x\geq 0$ creates a valid connection between different layers. In reality, logistic networks are relatively concentrated, but the solver tends to create a dense topology where infinitesimal flows exist on many arcs, complicating operational execution. Although a threshold-type constraint may avoid such circumstance, it is still difficult to determine specific values due to differences in business characteristics and experience. (2) Constraint Heterogeneity preventing Decomposition: Conventional decomposition methods like Column Generation (CG) rely on a solvable pricing sub-problem. In our network, the specific business constraints \eqref{eq:custom_constraints} are highly customized for different commodity subsets. Constructing a unified mathematical pricing oracle to capture these diverse rules is computationally inefficient. (3) Structural Rigidity: A minor change in network topology or a new business rule requires reconstructing the global constraint matrix $\mathbf{A}$ in \eqref{eq:custom_constraints}, making the system incapable of rapid iteration.

\vspace{-.4cm}
\begin{figure}[H]
    \centering
    \includegraphics[width=.9\linewidth, trim=0cm 1cm 0cm .8cm,clip]{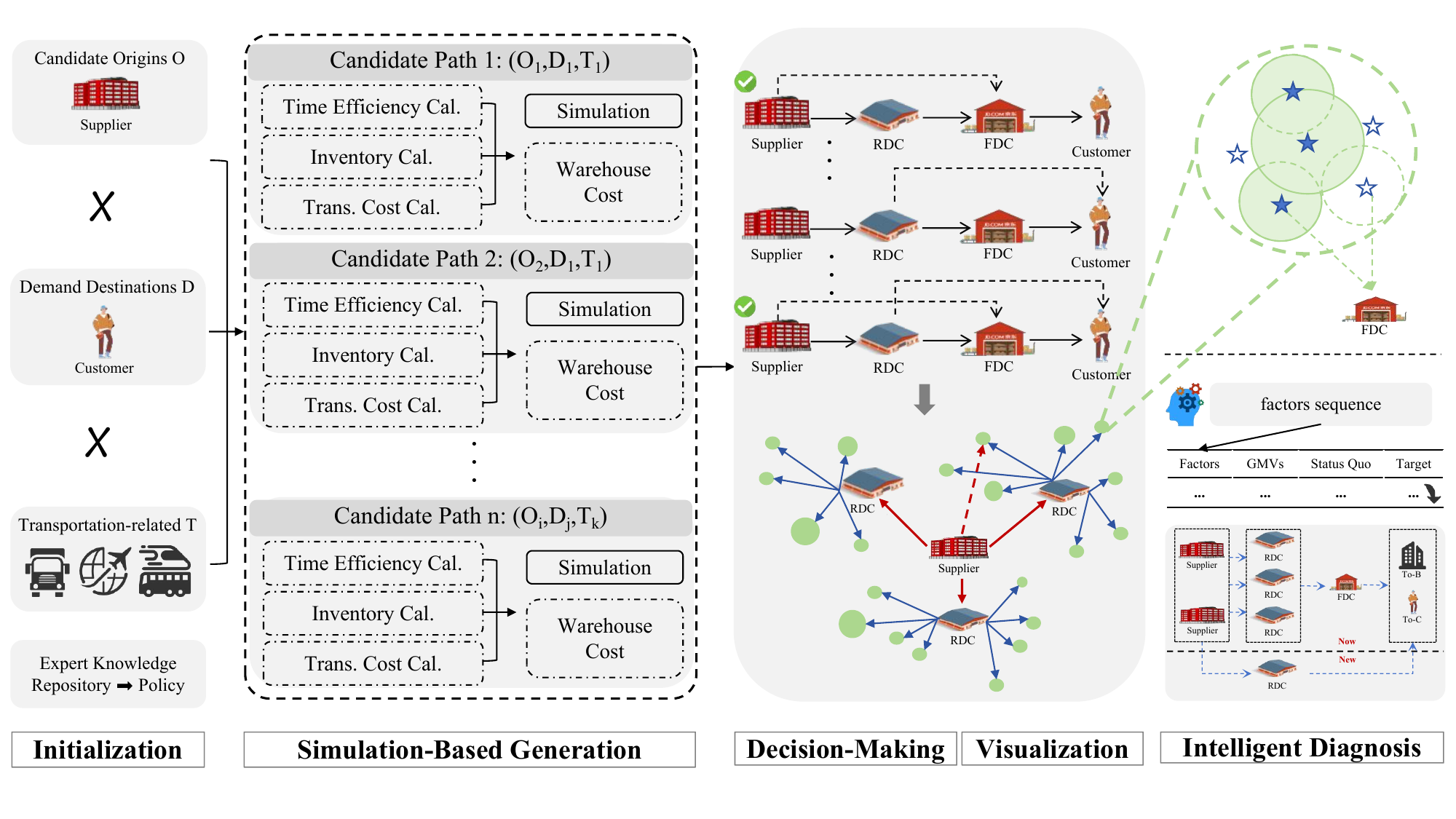}
    \caption{The Pipeline of SPORD (illustrated by the case of TNP)}
\label{fig:pipeline_SPORD}
\end{figure}
\vspace{-1cm}

To overcome these fundamental limitations across supply chain planning, we propose a paradigm shift from bespoke arc-based or state-based optimization to a unified path-based formulation (with nuanced but simple, basic models for different problems). Rather than asking the optimizer to construct a valid spatial arc or temporal state-transition while satisfying complex rules, SPORD massively generates operationally valid candidate paths before solving the model. The candidate paths satisfy local business rules during simulation, while global feasibility still depends on the subsequent decision model (e.g. different paths may share the same capacitated warehouse, etc.). The whole framework is engineered as a modular, five-stage systematic architecture, as shown in Figure \ref{fig:pipeline_SPORD}. The following sections introduce the process in detail.

\subsection{Simulation-Based Path Construction}

To address the challenge of balancing high-fidelity modeling with computational solvability (Challenge I), the SPORD method begins with a System Initialization. This layer functions as a high-resolution digital version of the real supply chain, ingesting heterogeneous business data to construct a structured description of the supply chain attributes based on network nodes and links. This layer preserves the granularity of individual operational entities, enabling the precise evaluation of Product $\times$ Node $\times$  Echelon $\times$  Transportation interactions.

\subsubsection{The Initialization: Defining Supply Chain Attributes.}
\label{initialization}

Formally, we define the supply chain network as a directed multigraph, $G=(\mathcal{N},\mathcal{{L}})$, where $\mathcal{N}$ represents the set of distinct logistical nodes (warehouses) and $\mathcal{L}$ denotes the set of varying transportation linkages (arcs). The node set is partitioned into subsets $\mathcal{N}=\mathcal{N}_{fac}\bigcup\mathcal{N}_{cust}$,  representing distinct facility types (e.g., suppliers, RDCs, FDCs) and customer demand points. Each node $n\in\mathcal{N}$ is characterized by a static attribute vector $\mathbf{a_n}$  (e.g., geospatial coordinates, facility type, processing capabilities, etc.). To capture the complexity of managing millions of SKUs, we define the product set $\mathcal{S}$. Each SKU $s\in\mathcal{S}$  is associated with a physical attribute vector $\phi_s=(vol_s,wgt_s,cat_s)$, denoting volume, weight, and handling category (e.g., fragile, cold-chain). These attributes are critical for the calculation of storage density and transportation fill rates. Demand is modeled as a discrete stream of orders $d=(s,n_{dest},t_{due},q)$, specifying the requested SKU $s$, delivery location $n_{dest}$, required delivery time $t_{due}$, and quantity $q$. Network connectivity is governed by four combinatorial relational modules that encode permissible inventory flow rules: (i) Production relations (source logic) defines the mapping $\mathcal{M}_{prod}:\mathcal{S}\rightarrow \mathcal{N}_{fac}$, specifying which supplier nodes constitute the sourcing origin for specific SKUs. (ii) Inventory relations (storage logic) defines the mapping $\mathcal{M}_{inv}:\mathcal{S}\rightarrow \mathcal{N}_{fac}$, determining the permissible storage nodes for each SKU category (e.g., hazardous materials must be stored in specialized chemical warehouses). (iii) Fulfillment flows (directional logic) defines the set of permissible logical arcs $\mathcal{A}_{logic}$, dictating the supporting relationships between nodes (e.g., Node $A$ is authorized to replenish Node $B$). This establishes the directed topology of network coverage. (iv) Transportation routes (physical attributes) defines the physical connection attributes $\mathcal{A}_{phys}$. For every logical arc, this module assigns detailed cost and time parameters, denoted as vector $\mathbf{w}_{ij}=(c_{trans},t_{lead})$. In our framework, routes filtered by these rules constitute the initial construction space for the simulation engine.

Besides, the Initialization Layer incorporates a comprehensive set of constraints and cost parameters, denoted as
$\Omega$,  which govern the feasibility and evaluation of logistical activities. $\Omega$ includes hard constraints (e.g., maximum warehouse throughput capacity $Cap_n$, strictly prohibited routes) and soft constraints (e.g., target delivery timeliness $\tau_{sla}$). We define a composite cost function $C(\cdot)$ that maps operational activities to financial metrics, including parameterized settings for warehousing costs (labor, etc.), transportation costs (fixed costs, distance/quantity-based variable costs), and inventory (holding) costs. To manage the search space effectively, \textit{NetSim} integrates a Human-in-the-Loop (HITL) mechanism. This mechanism allows domain experts to inject industrial know-how into the initialization process. This module effectively prunes the search space and aligns the model parameters with specific business contexts: (i) Context-Aware Node Strategies ($\mathcal{S}_{node}$). 
We introduce a tunable strategy vector $\mathcal{S}_{node}$ for facility nodes, enabling experts to toggle operational modes (e.g., steady-state, peak-surge, elastic expansion), dynamically adjusting effective capacity constraints $Cap_{n}(t)$ to reflect real-world resource scaling. (ii) Pruning of Routing Logic ($\mathcal{M}_{filter}$). To maintain tractability, \textit{NetSim} applies an expert-defined logical mask $\mathcal{M}_{filter}$ to pre-filter $\mathcal{A}_{logic}$ using business rules (e.g. \textit{product category $A$ implies direct-to-consumer flow}). The Initialization Layer discretizes the continuous flow space according to business-meaningful breakpoints: minimum order quantities per arc $(q_b)$, vehicle capacity fill-rate thresholds $(\eta)$, and expert-specified routing priorities, ensuring every instantiated path corresponds to a genuinely executable operational decision. (iii) Cost $\&$ Service Policy Parameterization ($\Pi_{policy}$). Cost and service parameters are policy-dependent. For example, experts select replenishment models (e.g., $(s,S)$, predictive demand-driven policies, or hybrid methods) for different SKUs, and configure region- and category-specific timeliness thresholds $\tau_{sla}(region,category)$, reflecting heterogeneous customer latency tolerances across the network. These configurations are institutionalized in a dynamic expert knowledge repository $\mathcal{R}_{exp}$, which maps historical problem instances to recommended parameter sets by scenario category (e.g., fresh product vs. electronics) and planning horizon (e.g., strategic vs. tactical). Each new instance is initialized from $\mathcal{R}_{exp}$, ensuring that the structural definition is not a cold start but a warm initialization with accumulated domain knowledge.

The Generation Layer then functions as a high-fidelity key performance indicator (KPI) oracle ($\mathcal{O}$), responsible for generating candidate paths and mapping a candidate topological structure as well as the input policy parameters into scalar attributes (e.g. cost $c_p$, service level $\alpha_p$) based on the aforementioned initialization. 
A critical design principle is comprehensive coverage: the Generation Layer simulates the complete set of operationally meaningful candidate paths rather than a sampled subset. At the intra-path level, flow discretization bounds the infinite continuous simplex to a finite set of business-valid configurations. At the inter-path level, expert pre-screening ($\mathcal{M}_{filter}$, Section \ref{initialization}) combined with JD.com's massively parallel computing infrastructure (detailed in Section \ref{sec:scalability}) enables the evaluation of tens of millions of candidate paths within operational time windows. This comprehensive coverage also addresses Challenge III directly: by simulating paths never historically exercised, SPORD transforms the sparse observational dataset into a dense, simulation-grounded cost landscape that supports rigorous counterfactual what-if analysis. To accommodate varying contexts of operational reality (static/ dynamic), we design a dual-mode mechanism: a static generation mode for macroscopic network planning, and a dynamic generation mode for microscopic inventory simulation.

\vspace{-0.3cm}
\subsubsection{Macroscopic Equilibrium: The Flow Balance Requirement.} 
\begin{figure}[H]
    \centering
    \includegraphics[width=0.65\linewidth]{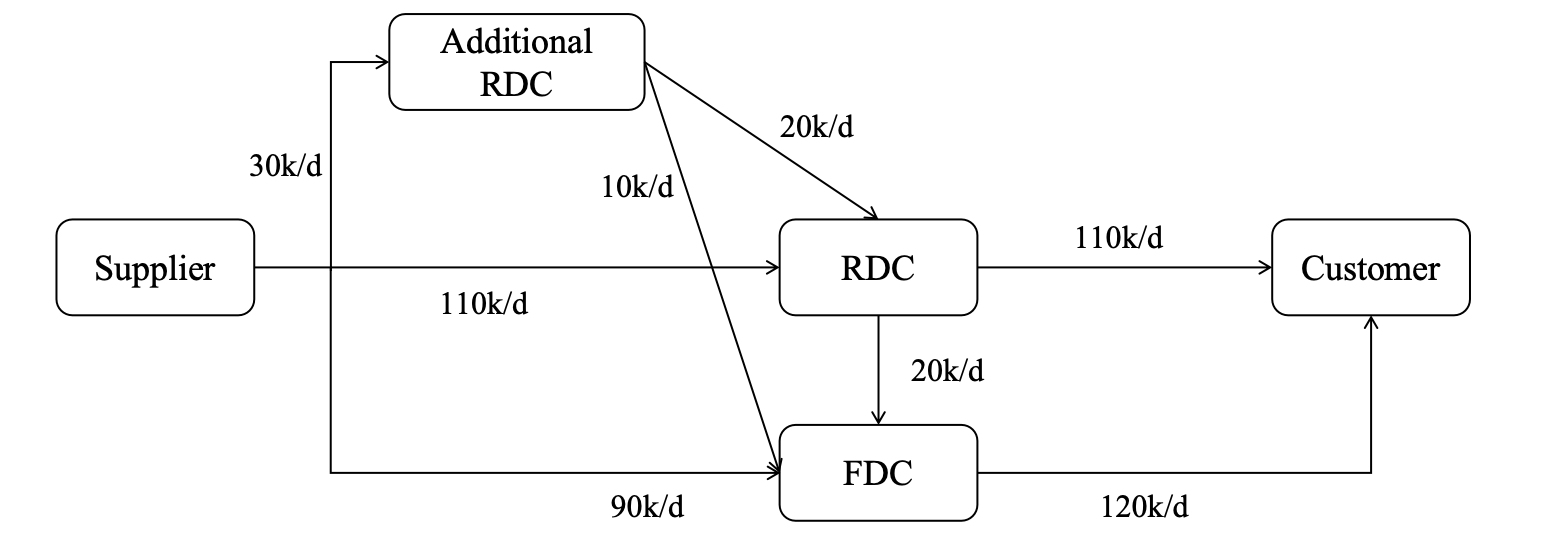}
    \caption{Illustration of Flow Balance Requirement for a Candidate Path}
    \label{fig:flow-balance}
\end{figure}

Building on the initialized network structure, the static generation mode constructs each candidate path by ensuring a macroscopic flow balance requirement: at every intermediate transshipment node, total inflow must equal total outflow, and the terminal demand sink must receive exactly the committed demand. Let $G_p=(N_p,A_p)$ be a subgraph extracted from the overall fulfillment network. As illustrated in Figure \ref{fig:flow-balance}, the isolated subgraph itself is a single candidate path $p$. It represents one potential routing configuration to fulfill the demand of Customer $D$ from Supplier $O$. The Generation Layer constructs millions of such candidate paths to fulfill all demands. For each path, the system operates in a conservation state where inflow equals outflow at every transshipment node $n\in N_p$. Let $d_{OD}$ be the demand for an Origin-Destination pair. The flow $f_{ij}$ on link $(i,j)$ is derived via backward propagation from the demand sink: $\sum_{k\in\delta^-(n)}f_{kn}=\sum_{j\in\delta^+(n)}f_{nj},\ \forall n\in N_p\backslash\{O,D\},$ $\sum_{k\in\delta^-(D)}f_{kD}=\sum_{j\in\delta^+(O)}f_{Oj}=d_{OD},$ where $\delta^-(n)$ means the in-degree node of $n$, and $\delta^+(n)$ means the out-degree node of $n$. Upon determining $f_{ij}$, the system selects the proper vehicle type $v^*$ and its corresponding quantity $m_{ij}$. This step explicitly couples transportation capacity with flow intensity, allowing for precise calculation of lead times  $L_{ij}$
and transport costs. Depending on the inventory strategy specified by experts (e.g., (s,S), base-stock, etc.), we then estimate the inventory cost along this path using the initial inventory and replenishment targets information.
\vspace{-0.5cm}
\vspace{-1.25cm}

\subsubsection{Microscopic Dynamics: Inventory Simulation.}
\label{micro dynamics}

While the macroscopic equilibrium provides a rapid static assessment, it masks the volatility of daily operations. To accurately evaluate the candidate paths under dynamic operational details for cases like WAP, we deploy the parallel track: the dynamic generation mode. This track models the microscopic temporal evolution of multi-echelon inventory systems over a discrete horizon $t\in\{1,\dots,T\}$.

\textbf{The Structure.} The dynamic system is characterized by its state vector at each period $t$, continuously perturbed by exogenous events, i.e. realized/estimated demand sequence, etc.
\begin{itemize}
    \item Inventory State: The vector $H_t=[I_t,IP_t,B_t]^T$ represents the system's \textit{memory}, encapsulating the on-hand inventory $I_t$, the inventory position $IP_t$ (including pipeline inventory), and the backlog quantity $B_t$.
    \item Exogenous Events: The vector $x_t=[D_t,\Lambda_t]^T$ represents external variables, where $D_t$ is the realized demand and $\Lambda_t$ indicates event triggers (e.g. review day arrival).
\end{itemize}

\textbf{State Transition.} The core of the dynamic simulation lies in its state transition logic, governed by predefined inventory control policies. Taking a periodic review $(R,S)$ policy as an illustrative example, the temporal evolution proceeds sequentially. 
\begin{itemize}
    \item Order Generation: On a designated review day, the system evaluates the current inventory position to determine the replenishment order quantity $O_t$, ensuring non-negativity: $O_t=\max\{0,S-IP_{t-1}\}\cdot\mathbb{I}(t\in\text{ReviewDays}).$
    \item System Update: The on-hand inventory $I_t$ evolves by integrating arrivals (orders placed $L$ days ago, where $L$ is the lead time) and demand subtraction, again processed to separate physical stock from backlogs: $I_t=(I_{t-1}+O_{t-L}-D_t)^+,$ $B_t=(D_t-(I_{t-1}+O_{t-L}))^+.$
\end{itemize}
The simulation generates a complete trajectory of inventory states by propagating these transition equations across the full horizon $T$. This trajectory is then aggregated to derive the cost attributes, $c_p$, which is the simulation outputs for the following optimization in the whole SPORD method.

\subsection{The Path Selection Model}
\label{sec:decision_making}

Following the path construction and high-fidelity simulation in the Generation Layer, the system produces a comprehensive set of candidate paths $\mathcal{P}$. Regardless of whether the problem context is static structural planning or dynamic warehouse assortment planning, each candidate $p\in\mathcal{P}$ is standardized and encapsulated as a self-contained proposal. It inherently carries a cost attribute $c_p$ and a resource consumption vector $a_p$. It is crucial to note that while the fundamental framework remains consistent across both static and dynamic domains, the specific mathematical structures of the base models do exhibit necessary variations depending on the problem types. But fortunately, because all complex and idiosyncratic operational details have been absorbed by the simulation phase, the final optimization model is reduced to a simple, fundamental, and highly interpretable Binary Integer Programming (BIP) formulation. Specifically, as established in Section \ref{grand_model}, the monolithic model embeds operational rules directly into the constraint matrix, making it rigid and computationally burdensome. SPORD removes this bottleneck. After the Generation Layer constructs and filters a large set of operationally valid candidate paths via parallel simulation, the remaining optimization problem reduces to a path-based formulation: given the precomputed operational costs and resource consumption attributes of each candidate path, the Decision-Making Layer selects a subset of candidates that covers demand and respects cardinality and capacity logic.

The model involves two classes of binary decision variables: one governing the selection of physical warehouse locations, and the other governing the selection of candidate paths—together encoding both the structural and the routing dimensions of the planning decision. Let $\mathcal{P}$ denote the set of all candidate paths generated by the simulation layer. Let $x_n \in \{0,1\}$ be the binary variable indicating whether candidate warehouse $n \in \mathcal{N}$ is selected, and $y_p \in \{0,1\}$ be the binary variable indicating whether candidate path $p \in \mathcal{P}$ is selected. Let $n_d \in \mathcal{N}_{demand}$ denote the demand point, $\mathcal{N}_{required}$ the set of required warehouses that must be chosen, and $\mathcal{N}_{banned}$ the set of banned warehouses that cannot be chosen. The candidate warehouse sets with cardinality limits denoted as $N_{card}$ and those with capacity limits denoted as $N_{cap}$. The optimization model is formulated as follows:
\vspace{-1cm}
\begin{subequations}
\setlength{\jot}{0pt}
\begin{align}
\label{TNP model}
    \min\  &C_s^T X + C_T^TY  \\
    \text{s.t.}\  & \sum_{p \in P:{n_d}\in p} y_p =1 , \quad   \forall n_d \in N_{demand}\label{demand-satis} \\
      &\sum_{p\in P:n \in p} y_p  \ge x_{n} ,\quad \forall n\in N \label{candi-path-relationship}\\
      & y_p \le x_n,  \quad\forall  p \in P,\ \forall n \in p\label{path-candi-relationship}
      \end{align}
      \begin{align}
      &K^{\min}_{card} \le \sum_{n\in N_{card}} x_n\le K^{\max}_{card} \label{cardinality-cons}\\
      &x_n=1 ,\quad\forall n\in N_{required}\label{required-cons}\\
      &x_n=0,\quad \forall n\in N_{banned}\label{banned-cons}\\
      &{C}^{min}_{cap} x_{n} \le \sum_{p\in P: n\in p} c^{cap}_p y_p, \forall n\in N_{cap} \label{capacity-cons-local}\\
      &{C}^{max}_{cap} x_{n} \ge \sum_{p\in P: n\in p} c^{cap}_p y_p, \forall n\in N_{cap} \label{capacity-cons-local-upper}\\
      &{C}^{min}_{SL} \le \sum_{p\in P} s_py_p \le {C}^{max}_{SL}\label{capacity-cons-global}\\
      &x_n \in \{ 0,1 \}, \quad\forall n \in N\label{candi-binary}\\
      &y_p \in \{ 0,1 \}. \quad\forall p \in P\label{path-binary}
\end{align}
\end{subequations}

\vspace{-1cm}
The objective is to minimize the total cost, including warehouse and transportation costs. Constraint \eqref{demand-satis} ensures that the demand of every customer point is satisfied. Constraint \eqref{candi-path-relationship} guarantees that if a candidate location is selected, at least one candidate path that includes this location must also be selected. Constraint \eqref{path-candi-relationship} enforces the reverse implication: if a candidate path is selected, all candidate locations that constitute that path must also be selected. Constraint \eqref{cardinality-cons} limits that for any subset of candidate locations with cardinality limits, the number of selected locations remains within the specified lower $K_{card}^{min}$ and upper $K_{card}^{max}$ bounds. Constraint \eqref{required-cons} requires that all warehouses in the set of required locations are selected. Constraint \eqref{banned-cons} ensures that no warehouse from the set of banned locations is selected.
Constraints \eqref{capacity-cons-local}, \eqref{capacity-cons-local-upper} and \eqref{capacity-cons-global} enforce local and global capacity limits on the candidate routes, respectively. For instance, for any $n\in N_{cap}$, the sum of the volumes of all chosen paths passing through the warehouse $n$ should be in the interval $[C_{cap}^{min}x_n,C_{cap}^{max}x_n]$.
Finally, the last two constraints define the domains of the decision variables for selecting candidate locations and candidate paths, typically as binary variables. This TNP formulation exemplifies the SPORD philosophy: complex operational realities (vehicle types, SKU attributes, time windows) are pre-evaluated in simulation, leaving the optimizer with a clean, binary selection task. For dynamic problems such as WAP, the Decision-Making Layer employs a structurally analogous but contextually adapted model, detailed in Online Appendix \ref{appendix:applications}.

The reduction from the monolithic model to the above model by SPORD yields two critical advantages: (i) interpretability and stability: the transparent BIP structure is readily auditable by business stakeholders even without advanced mathematical backgrounds. (ii) plug-and-play versatility: The base model is an adaptable, universal chassis. Swapping between different cases requires only a model chassis change, not a full re-engineering. The resulting sparse linear model is directly solvable by commercial solvers (e.g., Gurobi, CPLEX) without specialized decomposition.

\vspace{-.2cm}

\section{Scalability via Parallel Computing and List Scheduling}
\label{sec:scalability}
The mathematical simplicity of the Decision-Making Layer is achieved by offloading complex business constraints into the simulation. This decoupling allows standard solvers to solve the final model efficiently. Fundamentally, it shifts the computational burden upstream. The simulation engine must evaluate tens of millions of candidate paths and process intricate constraints. To ensure tractability, we leverage high-throughput parallel computing for independent path generation. For coupled orders with precedence constraints, we apply a scheduling algorithm.

\textbf{High-Throughput Simulation via Resilient Distributed Datasets (RDDs).} For the majority of candidate paths where flow dynamics are decoupled, we leverage the data parallelism inherent in the solution space. Unlike traditional disk-based \textit{MapReduce} paradigms, we utilize Apache Spark’s in-memory computing framework centered on Resilient Distributed Datasets (RDDs), which are immutable, fault-tolerant, and distributed collections of objects designed for parallel processing. The set of candidate paths $\mathcal{P}$ is encapsulated as a distributed RDD. The simulation oracle is broadcasted to worker nodes (i.e. machines or processes that actually perform the computational work). They receive tasks from a ``driver" program (which coordinates the overall execution) and process data, allowing the attribute computation (cost $c_p$, service metrics $\alpha_p$) to occur entirely in memory without incurring significant input/output overhead. This architecture minimizes latency by caching the Initialization Layer's structural data. In production environments, the system achieves extreme throughput and ensures the downstream Decision-Making Layer receives a comprehensive candidate space with negligible pre-processing time.

\begin{figure}[H]
    \centering
\includegraphics[width=.9\textwidth,trim=1cm 10cm 1cm 2cm,clip]{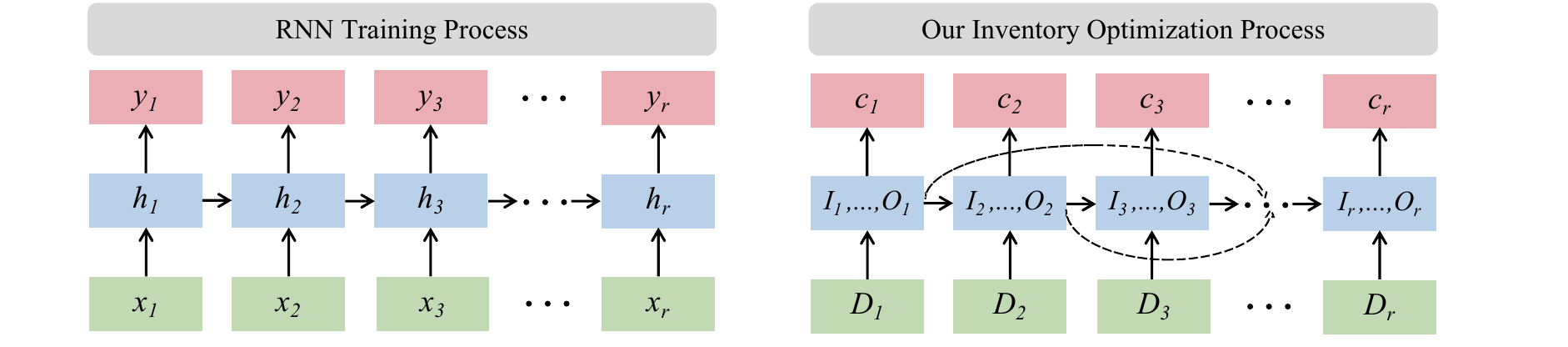}
    \caption{RNN-Isomorphic Inventory Simulation}
    \label{fig:RNN_inventory}
\end{figure}

\textbf{Acceleration via Matrix Vectorization and GPU Offloading.} For paths with dense temporal dynamics, serial processing creates a latency bottleneck. We address this via Matrix-Based Parallelism: the inventory simulation is isomorphic to a Recurrent Neural Network (RNN) forward process (Figure \ref{fig:RNN_inventory}), enabling decomposition into tensor operations offloaded to GPU.

The inventory simulation unfolds over a discrete time horizon $t\in\{1,\dots,T\}$. We map the core components of RNN cells to inventory dynamics as follows:
\begin{itemize}
    \item Hidden State ($h_t$) $\leftrightarrow$ Inventory State: The vector $\mathbf{H}_t=[I_t,IP_t,B_t]^T$ represents the system's \textit{memory}, encapsulating the physical on-hand inventory $I_t$, the inventory position $IP_t$ (including pipeline inventory), and the backlog quantity $B_t$.
    \item Input ($x_t$) $\leftrightarrow$ Exogenous Events: The vector $x_t=[D_t,\Lambda_t]^T$ represents external variables, where $D_t$ is the realized demand and $\Lambda_t$ indicates event triggers (e.g. review day arrival).
    \item Trainable Weights ($W$) $\leftrightarrow$ Policy Parameters: The static parameters governing the logic, such as the target stock level $S$ or reorder points $ROP$, correspond to the weights in an RNN. Our optimization goal is to find these parameters to minimize the loss function.
    \item Activation Function ($\sigma$) $\leftrightarrow$ Replenishment Logic: The control logic is mathematically equivalent to activation functions (e.g., the non-negativity of orders is inherently a ReLU operation).
\end{itemize}
\vspace{-0.5cm}
\vspace{-1cm}

The simulation executes a forward pass analogous to unrolling an RNN. At each time step $t$, the state updates based on the previous state $\mathbf{H}_{t-1}$, current input $\mathbf{x}_t$, and policy parameters $\Theta$:
\begin{itemize}
    \item Order Generation: Under a periodic review $(R,S)$ policy, the order quantity $O_t$ is computed via: $O_t=\text{ReLU}((S-IP_{t-1})\cdot \Lambda_t)=\max\{0,S-(I_{t-1}+\text{InTransit}_{t-1})\}\cdot\mathbb{I}(t\in\text{ReviewDays}).$
    \item System Update: The on-hand inventory $I_t$ evolves by integrating arrivals (orders placed $L$ days ago, where $L$ is lead time) and demand subtraction, again processed through a ReLU to separate physical stock from backlogs: $I_t=\text{ReLU}(I_{t-1}+O_{t-L}-D_t),$ $B_t=\text{ReLU}(D_t-(I_{t-1}+O_{t-L})).$
    \item Loss Function and Parameter Optimization: The objective function is the training loss $\mathcal{L}(\Theta)$, defined as the cumulative operational cost over the simulation horizon: $\mathcal{L}(\Theta)=\sum_{t=1}^T(h\cdot I_t(\Theta)+p\cdot B_t(\Theta)+K\cdot\mathbb{I}(O_t>0)),$ where $h$,$p$,$K$ denote holding, penalty, and fixed ordering costs. Standard back propagation and Grid Search or Gradient-Free Optimization can both be implemented to iteratively update $\Theta$. The system explores the parameter space (e.g. varying $S$ and $ROP$) to find the global minimum of the loss surface. The final minimized cost $\mathcal{L}(\Theta^*)$ is then extracted as the cost-related attribute $c_p$ for the subsequent Decision-Making Layer.
\end{itemize}

The inherent recurrent structure of our dynamic simulation is particularly amenable to massive parallelization, offering opportunities for accelerating computations through GPUs. For scenarios such as inventory simulation over a long horizon ($T=30$ days), the traditional iterative approach (e.g., for $t$ in $T$) is computationally inefficient due to the overhead of interpreter locks and lack of instruction-level parallelism. We restructure the operational logic by flattening the temporal and SKU dimensions. Instead of processing orders sequentially, we construct the state space as high-dimensional tensors (matrices). The simulation logic  $S_{t+1}=f(S_t,D_t,R_t)$ is decomposed into linear algebra operations. The inventory states, sales signals, and replenishment decisions for thousands of SKUs are aggregated into a unified matrix structure. These matrix-structured calculations are offloaded to CPU or GPUs for further acceleration. GPUs, with their thousands of cores, are architecturally optimized to perform these matrix manipulations in parallel, bypassing the bottlenecks of serial processing.

Speed, however, is only half the equation. The simulation pipeline outlined in Section \ref{micro dynamics} provides a theoretically sound architecture for state transitions. However, the Generation Layer must maintain strict logical fidelity to reliably evaluate costs (Challenge III). This fidelity refers to the degree to which its simulated trajectories accurately mirror the actual physical operations of the supply chain. We validate this fidelity via historical back-testing: replaying realized exogenous inputs (e.g. realized demand, etc.) through the engine and comparing simulated trajectories against ground-truth operational records. The results are shown in Section \ref{numerical-validation}.

\textbf{List Scheduling for Coupled Orders.} When multiple orders with precedence constraints (e.g., FIFO allocation) compete for the same SKU, standard parallel execution stalls. Naive blocking causes severe thread starvation. We resolve this via a list scheduling module that transforms the blocked execution model into an active scheduling, reducing makespan ($C_{max}$) from hours to minutes. Specifically, we model the execution as a Parallel Machine ($P_m$) Scheduling Problem with Precedence Constraints ($P_m|prec|C_{max}$). The solution involves constructing a Directed Acyclic Graph (DAG) and applying a critical-path algorithm for large-scale coupled cases.
\begin{itemize}
    \item Jobs ($\mathcal{J}$): The set of $\mathcal{N}$ coupled orders to be simulated. Each job $j$ has a processing cost $p_j$ (typically evaluated by processing time).
    \item Machines ($\mathcal{M}$): The finite set of available computing threads in CPU.
    \item Precedence Constraints ($\prec$): A strict partial order $i\prec j$ implies job $j$ cannot start until job $i$ releases the inventory lock on a shared SKU.
    \item Target: Minimizing makespan ($C_{max}$). The optimization goal is to minimize the maximum completion time across all resources.
\end{itemize}

We map the order stream to a Directed Acyclic Graph
$G=(V,E)$. A directed edge $(i,j)\in E$ is established if $i\prec j$. We then compute the \textit{height} $H_j$ for each node, representing the distance to the terminal state on the critical path. The priority $\pi_j$ for sorting is set to this \textit{height} value: $\pi_j=H_j=p_j+\max_{\{k|(j,k)\in E\}}H_k.$ Nodes with higher $H_j$ are prioritized as they represent the bottlenecks of the batch. The scheduler maintains a dynamic Ready Queue $Q_t$, containing tasks whose in-degree is zero (all predecessors completed): $\mathcal{Q}_t=\{j\in V|\text{deg}_{in}(j)=0\}.$ 

At any time step, an idle machine $k$ pulls the task $j^*$ with the highest priority: $j^*=\arg\max_{j\in\mathcal{Q}_t}\pi_j.$ Upon completion, $j^*$ is removed from $G$, instantly unlocking successor nodes for the next cycle, updating $\mathcal{Q}_t$. Empirical results demonstrate that this approach enables the processing of almost 500,000 coupled orders within minutes. The detailed illustration refers to Online Appendix \ref{appendix:list_scheduling}.

\vspace{-0.2cm}
\section{Industrial Applications}
\label{cases}
\vspace{-0.1cm}
Section \ref{sec:scalability} established the computational infrastructure (i.e. speed and fidelity) necessary for a credible planning engine. This section answers the more consequential question: does SPORD translate into better real-world decisions, and can those decisions be adopted by business organizations?

\vspace{-0.1cm}
\subsection{Numerical Validation}
\label{numerical-validation}
\vspace{-0.1cm}
We have introduced the matrix-vectorized parallel simulation and list scheduling algorithm as the computational backbone of SPORD in Section \ref{sec:scalability}. Now, we empirically validate the performance of our parallel technique. Using real-world SKU data (Replenishment/Sales/Inventory simulation for small/medium items over a 91-day horizon), we compare several computational methods: Traditional for-loops (two-level for loops), pandas.apply methods (a Python-native vectorized processing format) and our vectorized approach where all data is restructured into matrices for simultaneous computation. As illustrated in Table \ref{tab:Comparison_of_different_process_structure}, the transition from serial processing to matrix parallelism yields a dramatic performance leap. For a batch of 50,000 simulation instances, the serial method requires approximately 75.9 seconds. The matrix-parallel method completes the task in 2.1 seconds. This represents an over 30-fold increase in efficiency, and parameter optimization accelerated by GPUs can achieve drastically more improvement (Table \ref{tab:CPU_vs_GPU}). This allows the Generation Layer to handle computationally intensive simulations without slowing down the overall pipeline, ensuring that even the most complex cost attributes are generated in near-real-time.

\vspace{-0.5cm}
\begin{center}
\begin{table}[H]
\renewcommand{\arraystretch}{0.7}
\setlength{\tabcolsep}{6pt}
\caption{Time Consumptions of Different Calculation Strategies in CPU (10 cores \& 40 Threads)}
\label{tab:Comparison_of_different_process_structure}
\begin{center}
\scalebox{0.85}{
\begin{tabular}{cccc}
    \toprule
    \textbf{\symbol{35} of SKUs} & \textbf{for-loops} (s) & \textbf{pandas} (s) & \textbf{vectorization} (s)\\
    \toprule
   100 & 0.179184 & 0.165043 & 0.009640\\ 
   1000 & 1.570751 & 1.742290 & 0.026114\\
   10000 & 16.089718 & 16.234240 & 0.230672\\
   50000 & 75.988832 & 81.876154 & 2.109408\\
   100000 & 159.204360 & 165.418447 & 5.115564\\
   \toprule
\end{tabular}}
\end{center}
\end{table}
\end{center}

\vspace{-1.2cm}
\begin{center}
\begin{table}[H]
\renewcommand{\arraystretch}{0.7}
\setlength{\tabcolsep}{6pt}
\caption{Time Consumptions of CPU-vectorization vs. GPU-vectorization}
\label{tab:CPU_vs_GPU}
\begin{center}
\scalebox{0.85}{
\begin{tabular}{cccc}
    \toprule
    \textbf{\symbol{35} of SKUs+DCs} & \textbf{CPU} of 8 cores \& 32 Threads (s) & \textbf{GPU} of P40*1 (s)\\
    \toprule
   100 & 140 & 15\\
   1000 & 243 & 21\\
   10000 & 1071 & 71\\
   \toprule
\end{tabular}}
\end{center}
\end{table}
\end{center}

\vspace{-1cm}
The results in Tables \ref{tab:Comparison_of_different_process_structure} and \ref{tab:CPU_vs_GPU} confirm that the parallel techniques deliver order-of-magnitude throughput improvements that make exhaustive evaluation operationally feasible at JD.com's scale. 

Computational speed alone is insufficient. Considering that the pure algorithmic logic is frequently disrupted by idiosyncratic business behaviors (e.g., manual overrides, capacity adjustments), we adopt an incremental testing approach, progressively injecting operational complexities into the baseline model to guarantee simulation fidelity. The following progressive back-testing quantifies how closely \textit{NetSim}'s simulated inventory and sales trajectories track ground-truth operational records as progressively more real-world data is incorporated. Three metrics quantify this simulation-reality alignment. Let $I_{s,d,t}^{sim}$/$I_{s,d,t}^{act}$ and $S_{s,d,t}^{sim}$/$S_{s,d,t}^{act}$ denote simulated and actual inventory and sales for a specific SKU $s$ at Distribution Center (DC) $d$ on day $t$. The Cumulative Inventory Attainment Rate (CIA) measures the macroscopic volume alignment of the inventory, i.e., $CIA=\sum_{s,d,t}I_{s,d,t}^{sim}/\sum_{s,d,t}I_{s,d,t}^{act}$. The Cumulative Sales Attainment Rate (CSA) measures the macroscopic volume alignment of fulfillment capability, i.e., $CSA=\sum_{s,d,t}S_{s,d,t}^{sim}/\sum_{s,d,t}S_{s,d,t}^{act}$. The Granular Inventory Accuracy (GIA) is based on the Mean Absolute Percentage Error (MAPE), evaluating the absolute deviation at the most granular level, i.e., $GIA=1-\sum_{s,d,t}\big|I_{s,d,t}^{sim}-I_{s,d,t}^{act}\big|/\sum_{s,d,t}I_{s,d,t}^{act}$. With these metrics, we validate our simulation's accuracy against the real-world supply chains.  
 
\vspace{-.5cm}
\begin{center}
\begin{table}[H]
\renewcommand{\arraystretch}{0.7}
\setlength{\tabcolsep}{6pt}
\caption{Progressive Enhancement of Simulation Logical Fidelity}
\label{tab:Progressive Enhancement of Simulation logical fidelity}
\begin{center}
\scalebox{0.85}{
\begin{tabular}{ccccc}
    \toprule
    \textbf{Scenario Evolution} & \textbf{CIA}  & \textbf{CSA} & \textbf{GIA} & \textbf{Cumulative Simulation Accuracy}\\
    \toprule
   Baseline & 73.4\%	 & 75.3\%	 & 51.1\%	 & 74.3\%\\
    + Initial In-Transit Inventory	 & 78.3\%	 & 77.2\%	 & 54.4\%	 & 77.7\%\\
    + Manual Order Interventions	 & 108.3\%	 & 96.6\% & 	80.6\%	 & 94.2\%\\
    + Fulfillment Capacity Adjustments	 & 94.9\%	 & 92.7\%	 & 82.1\%	 & 93.8\%\\
    + Banned-List Constraints & 	94.9\%	 & 92.6\%	 & 82.5\%	 & 93.7\%\\
    + Order Placement Adjustments	 & 95.1\% & 	92.6\% & 	82.3\% & 	93.9\%\\
    Full-Data Driven Simulation	 & 99.9\%	 & 99.9\%	 & 99.7\%	 & 99.9\%\\
   \toprule
\end{tabular}}
\end{center}
\end{table}
\end{center}

\vspace{-1cm}
As shown in Table \ref{tab:Progressive Enhancement of Simulation logical fidelity}, the naive baseline model, relying solely on standard replenishment logic, struggles to capture the reality, yielding a Granular Inventory Accuracy (GIA) of only 51.1\%. As we systematically incorporate real-world operational information, the fidelity metrics exhibit an upward trajectory. Notably, the introduction of manual orders (i.e. manual overrides data by business staff) corrects a massive variance in cumulative attainment, while fulfillment adjustments refine the granular accuracy to over 82\%. Ultimately, when driven by the full spectrum of historical operational data, the simulation achieves near-perfect alignment with reality, recording a cumulative simulation accuracy of 99.9\% and a granular SKU-DC-Day accuracy of 99.7\%. This exceptional logical fidelity confirms that the \textit{NetSim} platform is qualified to serve as a well-performed cost oracle for the subsequent Decision-Making Layer.

Together, the experiments provide quantitative confirmation that the parallel architecture delivers the throughput required for industrial-scale candidate evaluation, and the progressive fidelity validation confirms that \textit{NetSim} can serve as a reliable cost oracle grounded in operational reality.

\vspace{-0.1cm}
\subsection{Implementation}
\label{implementation}
\vspace{-0.1cm}
To demonstrate how the SPORD method achieves universal applicability across diverse SCP domains, the following subsections detail its deployment at two ends of the spectrum: static TNP (Section \ref{TNP_application}) and dynamic WAP (Section \ref{IA_application}). 

\vspace{-0.3cm}
\subsubsection{Application Scenario I: Trunk Network Planning.}
\label{TNP_application}

\textit{NetSim} simulates millions of candidate paths. A path $p$ is defined as a tuple $(O_i,D_j,T_k)$, representing a complete logistical chain from Origin $O_i$ to Destination $D_j$ using Transportation Mode $T_k$. Detailed operational parameters such as the inventory management policies, and further transportation specifics, are encoded as attributes, i.e. part of the $T_k$ parameter of each path. During generation, \textit{NetSim} strictly filters paths against local constraints, such as specific routing rules, etc. It then pre-calculates the operational cost $C_s^T$ and transportation cost $C_T^T$ for each \textit{path}, incorporating elements like transportation time, etc. The problem is finally abstracted into an integer program (refer to Section \ref{sec:decision_making}) to select the optimal subset of pre-validated paths that minimize the total cost while satisfying generic constraints. By moving business logic into the simulation-based Generation Layer, the solver focuses purely on the selection, reducing solution time for large-scale instances (up to 80 million candidate paths before implementing dimension reduction) to nearly 90 seconds.

\vspace{-0.3cm}
\subsubsection{Application Scenario II: Warehouse Assortment Planning.}
\label{IA_application}

In each decision period $t$ within the planning horizon, for a candidate SKU-warehouse pair $(i,o)$, \textit{NetSim} executes a detailed simulation process for a preset time shot (a trajectory until $T$, e.g. 30-days plan). This engine systematically updates the state of the inventory system by tracking on-hand stock, pipeline inventory, arriving inventory, new order placements, and the amount of demand that can be successfully fulfilled. Based on these updated states, it then calculates the end-of-period inventory and the costs incurred during that specific period, which are composed of holding costs, transportation costs, and the lost sales along the trajectory. At the end of each period's simulation, a cumulative total cost from the start of period $t$ up to the preset time horizon is computed based on the functions defined in Online Appendix \ref{appendix:applications}. Then, this cumulative total cost is incorporated as the objective function. By simulating trajectories, \textit{NetSim} dynamically identifies whether an item should be positioned in a RDC or a FDC, balancing the Endless Aisle (massive SKUs) availability with fulfillment costs. From 2025 to the present, JD.com has achieved a year-on-year reduction of over 73 million dollars in supply chain costs attributable to WAP applications.

\vspace{-0.3cm}
\subsubsection{Visualization \& Verification: Bridging the Cognitive Gap.}
\label{sec:visual}

\begin{figure}[H]
    \centering
\includegraphics[width=.7\textwidth,trim=0cm 1cm 0cm 1cm,clip]{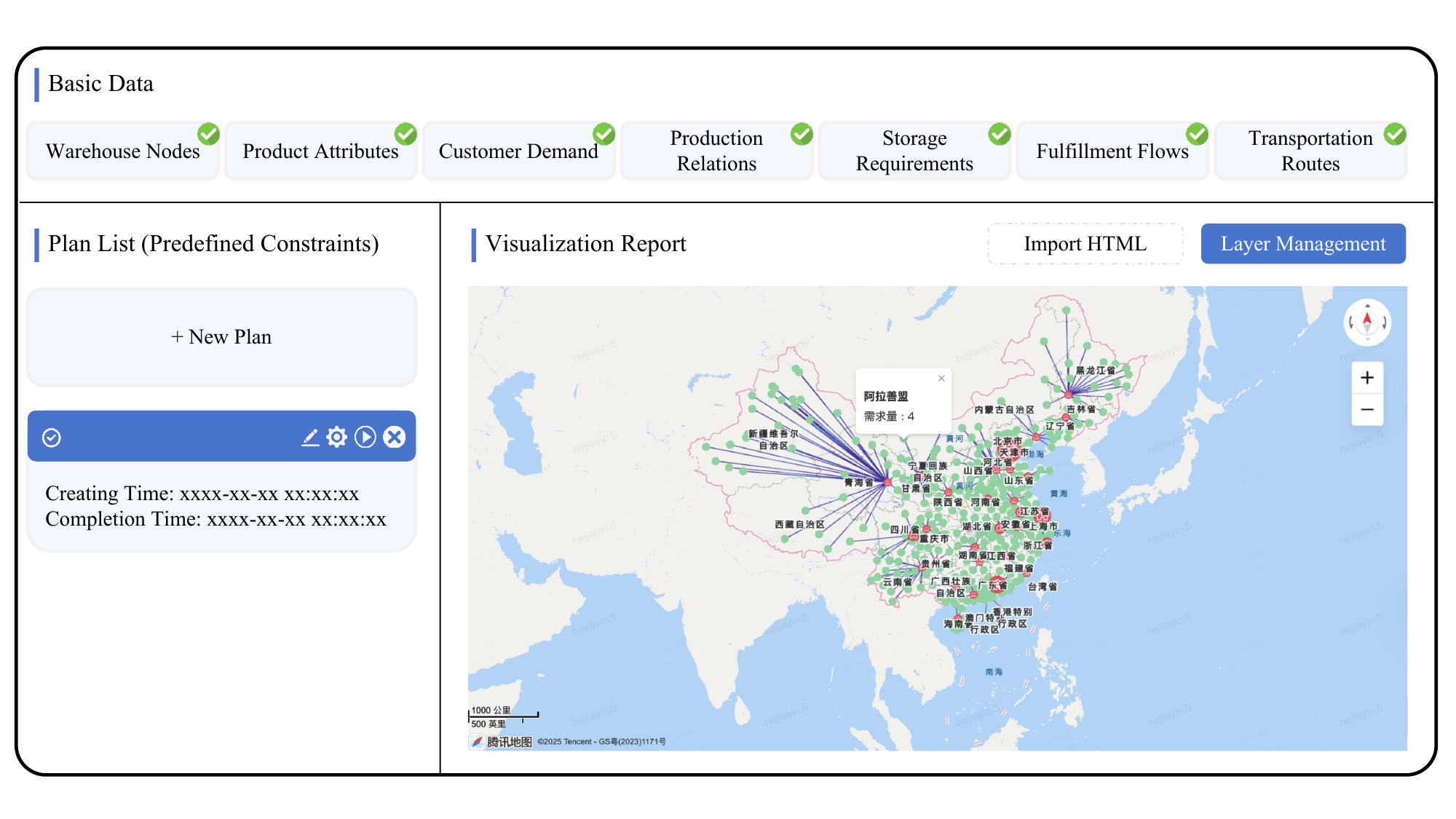}
    \caption{The Schematic Interface of NetSim in JD.com (illustrated by a TNP solution)}
    \label{fig:NetSim_sur}
\end{figure}

During the early piloting phases of \textit{NetSim}, JD.com's retail partners and regional logistics directors consistently raised a practical concern: they were extremely hesitant to adopt a static mathematical blueprint. From their frontline perspective, the e-commerce landscape is inherently volatile, and they requested a mechanism to observe how the recommended network would behave under unexpected stress before signing off on million-dollar structural changes (Challenge II). To consider this business feedback, our interactive module empowers the domain experts or stakeholders to inject perturbation scenarios (e.g., a sudden 20\% demand surge in a specific region). The system then computes the optimal solution for the perturbed state, and generates a set of adjacent solutions. We define the solution stability metric to quantify the robustness of the network design. By visualizing the initial solution alongside the perturbed one and their respective KPIs, decision-makers transition from passive recipients of an algorithm to active participants in the design process. They can visually verify whether the proposed solution shatters under volatility or gracefully absorbs the shock. This mechanism ensures that the final deployed solution is not merely a mathematical optimum for a deterministic snapshot, but a resilient configuration validated by domain expertise, thus also promoting the progress of adoption. In Figure \ref{fig:NetSim_sur}, the layer aggregates selected path flows at each facility and transport leg, translating mathematical outputs into concrete operational attributes: node attributes (e.g., facility throughput, capacity utilization) and edge attributes (e.g., freight volume). Executives can thereby assess macroscopic structure and micro-operational bottlenecks at a glance.

\vspace{-.5cm}

\vspace{-1.2cm}
\subsubsection{Intelligent Diagnosis: Closing the Decision Loop.}
\label{sec:intelligent}

Field feedback from JD.com's frontline warehouse managers and third-party merchants further exposed a post-implementation challenge: operational drift. They reported that over a planning cycle of several months, the actual empirical performance (e.g. fulfillment costs, cross-regional delivery rates) gradually diverges from the simulated baseline due to micro-level execution variances, shifting consumer behaviors, or localized capacity bottlenecks that emerge after the initial planning phase. To transform the SPORD framework from a static planning tool into a self-correcting dynamic system, we have designed an automated diagnostic feedback loop, which is the fifth step of our SPORD method. Conceptually analogous to a policy iteration mechanism, the system continuously monitors post-deployment KPIs against the simulated ground truth. When empirical metrics (e.g., fulfillment costs or lead times) deviate significantly from the standards given before, a hierarchical attribution tree decomposes the variance across intertwined operational factors. By applying a statistical pruning technique to isolate the root causes, the system quantifies the theoretical improvement gap weighted by actual business volume. These quantified anomalies are then translated into structural constraints or policy parameter adjustments, which can be fed back into the initialization for the subsequent planning epoch. This closed-loop architecture ensures sustained systemic resilience and continuous optimization. We illustrate the process with a real case from JD.com in Online Appendix \ref{appendix:intel_diagnosis}.

\subsection{Field Impact}
\label{field-impact}

Since \textit{NetSim}'s deployment in 2025, the platform has optimized end-to-end supply chain decisions for over 20,000 suppliers across JD.com's national network. The majority of these are small and medium-sized enterprises whose pre-\textit{NetSim} planning amounted to a regional manager's intuition, a spreadsheet calibrated once and never revisited, or simply last season's assignment carried forward. For the first time, their assortment and routing decisions are evaluated against a simulation of their actual supply chain, real demand patterns, real lead times, real replenishment constraints, rather than a planner's best guess. The verified outcomes across multiple live planning cycles are significant. For TNP, the cross-regional fulfillment rate dropped from 6.1\% to 4.9\%, a 20\% reduction; at JD.com's order volumes, it translates directly to faster deliveries and lower freight spend at national scale. For WAP, year-on-year supply chain cost savings attributable to WAP optimization have exceeded 73 million dollars annually. The platform's verified monthly carbon reduction of approximately 5,745 tCO2e is therefore what better decisions look like at scale: fewer unnecessary long-haul transshipments, less inventory redundancy, more precise pre-positioning in place of expedited shipping.

To provide a finer-grained view of this impact, we conduct a cost optimization analysis spanning January 2025 through June 2026 across ten major business units: the Large Supermarket Group, Home Appliances \& Furniture Group, 3C Digital Group, Fashion \& Apparel Group, Automotive Division, JD Health, Government \& Enterprise Division, Innovative Retail Group, Private Brand Division, and JD Industrial. For each unit, the direct fulfillment cost under the SPORD-recommended plan is benchmarked against the incumbent baseline on a monthly basis, and year-on-year improvements are computed for each of the first six months of 2026. The eighteen-month record reveals a consistent pattern of deepening optimization. Units onboarded in earlier rollout phases exhibit progressively larger absolute savings as the platform accumulates planning cycles and refines its candidate path space. The 3C Digital Group exemplifies this trajectory: year-on-year monthly savings grew from approximately 0.73 million dollars in January 2026 to 1.72 million dollars by June 2026—more than doubling within a single half-year. JD Industrial displays an even steeper acceleration, achieving a nearly 27-fold increase over six months, reflecting the compounding effect of route-level learning as more supplier-SKU combinations become eligible for SPORD-governed direct delivery. These trajectories confirm that SPORD's value accrual is not front-loaded: each newly onboarded product category or logistics corridor becomes an additional source of margin recovery without requiring bespoke re-engineering.

Viewed across business units, the performance snapshot illustrates both the breadth and heterogeneity of \textit{NetSim}'s impact. The Large Supermarket Group registers the single largest absolute improvement—approximately 11.16 million dollars in year-on-year savings in June 2026 alone, reflecting the outsized direct-delivery volumes of this unit and the correspondingly high leverage that route-level optimization yields at scale. The Fashion \& Apparel Group records more moderate but consistently positive savings (0.37 million dollars in June 2026), consistent with the category's lower average order value and more geographically concentrated fulfillment patterns. The Automotive Division and JD Health, both relatively newer to direct-delivery optimization, show steady positive trajectories (0.10 million dollars and 0.22 million dollars in June 2026, respectively), underscoring SPORD's generalizability to verticals with highly idiosyncratic handling and routing constraints. Although a small number of unit-month observations show negative year-on-year values, most notably the Large Supermarket Group in January 2026 (-1.14 million dollars) and marginal negatives in the Private Brand Division, they reflect a structural feature of progressive network expansion: as new product categories and logistics lanes are brought under SPORD-governed planning, they initially interact with adjacent flows (e.g. sharing warehouse capacity, transportation slots, and replenishment cycles) previously optimized independently. During the transition window, some cross-unit interactions manifest as transient cost increases for individual units even as the network-level trajectory remains favorable. Crucially, recovery follows in every case. The Large Supermarket Group makes this point most forcefully: after the -1.14 million dollars observation in January 2026, it recovers to 12.14 million dollars in February 2026 and then sustains positive.

\vspace{-0.4cm}
\section{Discussion}
\label{analysis}

\vspace{-0.1cm}
The large-scale deployment of the \textit{NetSim} platform at JD.com yields insights that extend far beyond specific algorithmic performance metrics. In this section, we summarize these empirical observations into broader managerial and architectural insights for the supply chain community.

\vspace{-0.2cm}
\subsection{Philosophy: Decoupling, Human-Algorithm Interaction, and Isomorphism}
\label{isomorphism}

\vspace{-0.1cm}
SPORD advances the SCP literature along three dimensions. \textit{Fidelity and generalizability}: standardizing static routing and dynamic inventory trajectories into a universal representation eliminates the need for bespoke models across business domains. \textit{Scalability and speed}: massively parallel simulation and GPU-accelerated matrix operations transform combinatorial explosions into tractable selection tasks, enabling near real-time decision cycles. \textit{Trust and continuous improvement}: the transparent candidate-path mechanism, HITL verification, and Intelligent Diagnosis feedback loop convert algorithmic outputs into auditable recommendations, sustaining stakeholder buy-in across planning epochs. These achievements rest on three design principles. \textit{Decoupling}: separating feasibility evaluation from prescriptive selection minimizes the blast radius of operational updates. Most parameter changes are absorbed non-invasively by the initialization, leaving the solver untouched. \textit{Human-algorithm symbiosis}: by encoding expert know-how into the knowledge repository, the system converts tacit individual experience into durable organizational assets, enabling warm-start initialization across planning cycles. \textit{Computational isomorphism}: conceptualizing inventory dynamics as an RNN-equivalent computation graph allows discrete logistical events to be mapped into high-dimensional tensors, unlocking massive GPU parallelism. Consequently, rather than conducting heuristic trial-and-error, \textit{NetSim} evaluates millions of candidate paths simultaneously, mapping a high-fidelity cost landscape from which solvers can instantly extract global optima.

\vspace{-0.2cm}
\subsection{Granularity Paradox: Theoretical Precision vs. Operational Feasibility}

\vspace{-0.1cm}
Our practical deployment has illuminated a critical determinant of algorithmic adoption: the calibration of decision granularity. 
Researchers naturally gravitate toward finer resolution to maximize theoretical optimality. However, we observe that the solution validity depends not only on mathematical precision but on alignment with the organization's operational execution bandwidth. For instance, while algorithms can generate precise, atomic-level instructions for specific orders, real-world logistical management often operates at coarser aggregates, such as facility-level resource allocation or route-level scheduling. Hyper-granular solutions, therefore, face a dual challenge: they are often cognitively difficult for stakeholders to accept and logistically prohibitive to execute. Beyond the explicit constraints encoded within the model, a vast space of tacit operational frictions governs day-to-day execution. Accordingly, SPORD constrains decision granularity to match the execution horizon of the business environment, acknowledging that a robustly implementable near-optimal solution consistently outperforms an unimplementable global optimum.

\vspace{-0.2cm}
\subsection{Future Extensions}
\label{conclusions}
\vspace{-0.1cm}
Three directions emerge as most consequential for future research. First, \textit{NetSim} already functions as a high-fidelity digital replica of JD.com's physical operations; the natural next step is to elevate it from a planning tool into a generative world model for supply chains, one that autonomously generates and reasons over hypothetical operational scenarios, internalizing the causal structure of the network well enough to simulate any intervention forward. Second, the current pipeline treats simulation and optimization as sequential; a richer OR–AI integration would close the loop: LLMs or planning agents generate strategic hypotheses (``what if we consolidate three regional DCs into a hub-and-spoke cluster?"), SPORD instantly simulates and costs them, and the optimizer's selection decisions in turn guide the next round of hypothesis generation—creating a human-AI-OR reasoning cycle in which expert knowledge, generative intelligence, and rigorous combinatorial optimization each play an irreplaceable role. Third, one persistent barrier to progress in supply chain optimization research is the scarcity of large-scale, operationally grounded benchmarks; most published results are validated on stylized or proprietary datasets that resist reproduction. We are exploring the possibility of partially opening the \textit{NetSim} platform to the research community so that researchers have a shared, industrially realistic testbed on which to develop and compare new methods. 

\vspace{-0.2cm}
\bibliographystyle{informs2014}
\bibliography{ref}

\end{document}


\ECSwitch  

\begin{center}
  {\Large Online Supplement}\\
  {\Large ``SPORD: A Simulation-Propose-then-OR-Dispose Approach for Supply Chain Planning"}
\end{center}
\bigskip
\section{Application Scenario Model (for WAP)}
\label{appendix:applications}

In dynamic WAP, traditional approaches face a fundamental difficulty: assortment decisions (which SKUs to stock at which facilities) are intrinsically intertwined with temporal inventory evolution and replenishment policies. When these components are encoded directly into a single dynamic programming model, the resulting formulation quickly becomes intractable at operational scale, and counterfactual evaluation is further limited by the reliance on historical decision records. Under SPORD, this coupling is handled by construction: the Generation Layer simulates the resulting SKU-to-warehouse temporal trajectories under the specified inventory policies and operational boundary conditions, producing standardized candidate outcomes whose cost and service implications are evaluated by simulation. Consequently, the OR-Dispose stage here does not re-simulate inventory dynamics; it instead solves a reduced selection formulation over the simulated candidate trajectories, using their simulation-derived costs and feasibility information to produce an implementable assortment plan that satisfies coverage and operational constraints. The resulting mathematical model in Eqs.\eqref{WAP_model}-\eqref{addition-cons} therefore represents the basic decision structure for WAP through SPORD, translating simulation outputs into actionable assortment decisions. During the process of dynamic simulation, the simulation engine systematically updates the state of the inventory system by tracking on-hand stock $I_{i,o,d,t}$, pipeline inventory $H_{i,o,d,t}$, arriving inventory $P_{i,o,d,t}$, new order placements $O_{i,o,d,t}$, and the amount of demand that can be successfully fulfilled $S_{i,o,d,t}$. Based on these updated states, it then calculates the end-of-period inventory $I_{i,o,d,t+1}$ and the costs incurred during that specific period, which are composed of holding costs $CI_{i,o,d,t}$, transportation costs $CF_{i,o,d,t}$, and the lost sales $CL_{i,o,d,t}$ along the trajectory. At the end of each period's simulation, a cumulative total cost $TC_{i,o,d}$ from the start of period $t$ up to the preset time horizon is computed.
\vspace{-.2cm}
\begin{subequations}
\setlength{\jot}{0pt}
\begin{align}
\label{WAP_model}
    \qquad\qquad \min \  &\sum_{i,o,d}\quad TC_{i,o,d}x_{i,o,d}\\
    \text{s.t.} \qquad  &\sum_{o}x_{i,o,d} =1,\  \forall (i,d) \ \text{existing} \label{demand-satisfaction}\\
      &x_{i,o,d} \le y_{i,o},\  \forall i,o,d \label{x-y}\\
     &\sum_{d}x_{i,o,d} \ge y_{i,o},\  \forall i,o \label{efficiency-y}\\
     &\sum_{i,o,d} CX_{i,o,d}x_{i,o,d} \le \epsilon. \label{addition-cons}
\end{align}
\end{subequations}

\noindent Constraint \eqref{demand-satisfaction} ensures that all demands are covered. Constraint \eqref{x-y} ensures that if product/SKU $i$ is not stocked at the distribution center $o$ (i.e. $y_{i,o}=0$), then the customer $d$ won't be satisfied by the $(i-o-d)$ (i.e. $x_{i,o,d}=0$). Constraint \eqref{efficiency-y} guarantees that every selected warehouse should at least cover 1 demand, otherwise the candidate warehouse won't be chosen. The last constraint \eqref{addition-cons} corresponds to additional business requirements, where the $CX_{i,o,d}$ can refer to other metrics, e.g., specific cost budgets, inventory turnover ratios for designated routes, and quotas allocated to particular distribution lines. The total cost of decision $x_{i,o,d}$ over the planning horizon is $TC_{i,o,d} =\sum_{t}^T\{CI_{i,o,d,t}+CF_{i,o,d,t}+CL_{i,o,d,t}\}$, calculated by:
\vspace{-.2cm}
\begin{subequations}
\setlength{\jot}{0pt}
\begin{align}
   \qquad\qquad\qquad  &I_{i,o,d,t} = I_{i,o,d,t-1} + P_{i,o,d,t},\\
    &H_{i,o,d,t} = H_{i,o,d,t-1} - P_{i,o,d,t},\\
    &O_{i,o,d,t} = -min\left\{I_{i,o,d,t} + H_{i,o,d,t} \right. \nonumber\\
    &\qquad- TI_{i,o,d,t}, 0\} * R_{i,o,d,t},\\
    &H_{i,o,d,t} = H_{i,o,d,t} + O_{i,o,d,t},\\
    &P_{i,o,d,t+VLT_{i,o,d,t}} = O_{i,o,d,t},\\
    &S_{i,o,d,t} = min\{I_{i,o,d,t}, D_{i,o,d,t}\},\\
    &I_{i,o,d,t+1} = I_{i,o,d,t} - S_{i,o,d,t},\\
    &CI_{i,o,d,t} = hc_i * I_{i,o,d,t},\\
    &CF_{i,o,d,t} = fc_{i,o,d} * S_{i,o,d,t},\\
    &CL_{i,o,d,t} = lc_i * (D_{i,o,d,t}-S_{i,o,d,t}).
\end{align}
\end{subequations}

\section{The List Scheduling Algorithm}
\label{appendix:list_scheduling}

As outlined in Section 4, when multiple orders 
compete for the same SKU inventory pool under a precedence allocation 
discipline, their simulation cannot proceed independently: a downstream order is not permitted to claim inventory until all higher-priority upstream orders sharing that SKU have been fully processed and their inventory locks released. Pure parallel execution cannot resolve these coupled resource constraints. We therefore introduce a List Scheduling algorithm to bridge this operational gap. This algorithm functions as a capacity allocator within the Generation Layer. It extracts conflicting fulfillment requests and processes them sequentially based on predefined priority rules. By doing so, the algorithm ensures strictly feasible resource assignments during the simulation phase.

\begin{figure}[H]
    \centering
    \includegraphics[width=\textwidth,trim=1.5cm 0cm 2.5cm 1.5cm,clip]{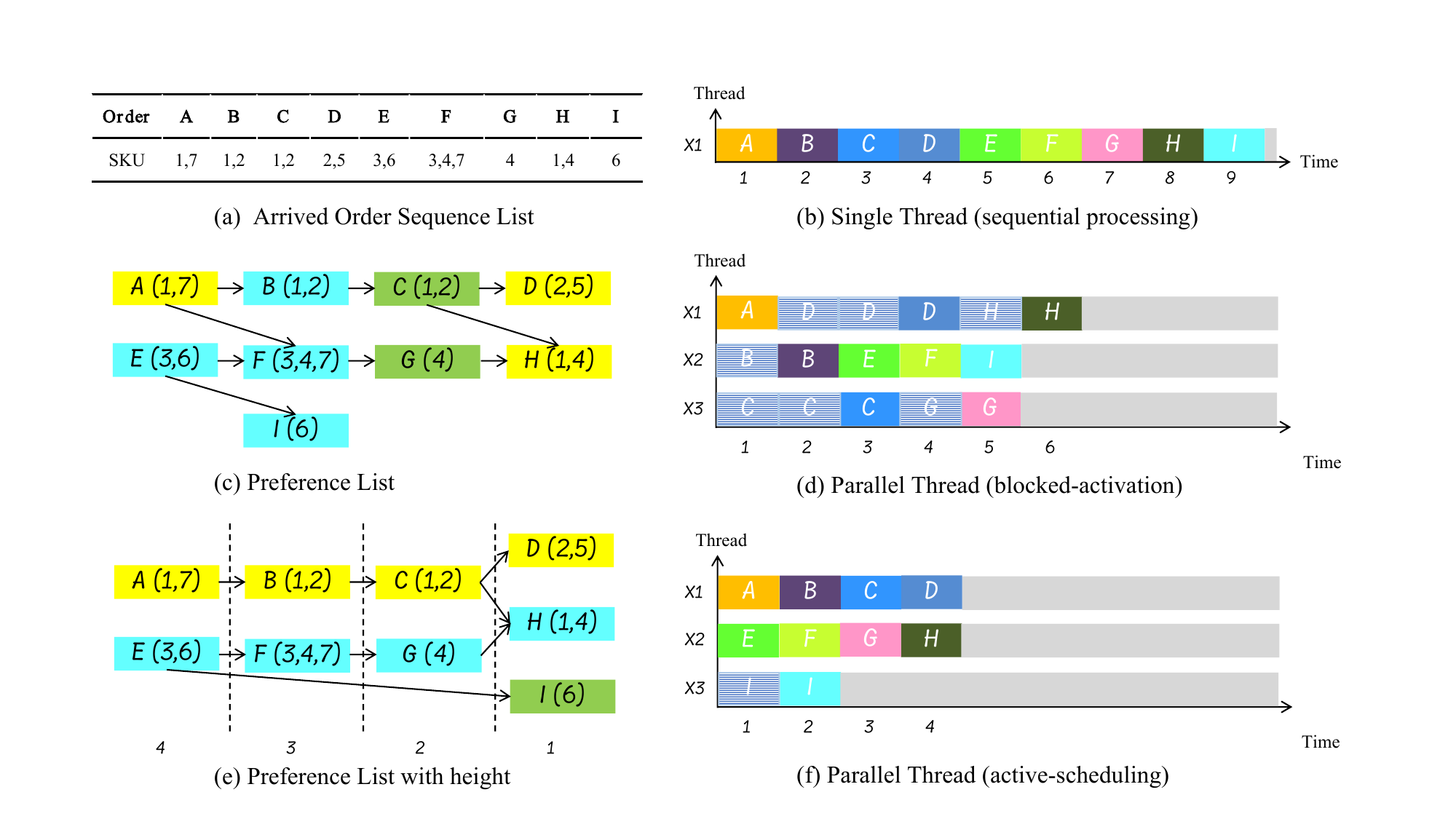}
    \caption{Illustration of List Scheduling Algorithm}
    \label{fig:list_scheduling}
\end{figure}

Figure \ref{fig:list_scheduling}(a) presents the given arrival sequence of orders. The input to the algorithm is a stream of $N$ orders, each associated with one or more SKUs it competes for. In the illustrative example, nine orders $\{A, B, C, D, E, F, G, H, I\}$ arrive, with their respective SKU dependencies listed alongside (e.g., Order~$A$ involves SKUs~$\{1,7\}$, Order~$B$ involves SKUs~$\{1,2\}$, etc.). This arrival sequence defines the raw input to the scheduler, but the arrival order itself does not determine execution priority---that is instead derived from the structural dependency analysis conducted in subsequent steps.

The core structural insight of the algorithm is that SKU-level conflicts define a partial order over jobs: if orders $i$ and $j$ share a common SKU and $i$ must be processed before $j$, we establish a directed edge $i \prec j$. Figure \ref{fig:list_scheduling}(c) depicts the resulting \emph{Directed Acyclic Graph} (DAG) $G = (V, E)$, where nodes represent orders and directed edges encode these precedence constraints. For instance, since Orders~$A$, $B$, $C$, and $H$ all involve SKU~$1$, they form a dependency chain; similarly, Orders~$E$, $F$, $G$ are linked through SKUs~$\{3,4\}$. This DAG captures the full dependency structure of the order batch and serves as the foundation for all subsequent scheduling decisions.

Figure \ref{fig:list_scheduling}(e) augments the DAG from Figure \ref{fig:list_scheduling}(c) with \emph{node heights} $H_j$, computed via a reverse topological traversal. The height of a node $j$ represents the total processing time along the longest downstream path from $j$ to any terminal node:
\[
    H_j = p_j + \max_{\{k \mid (j,k) \in E\}} H_k,
\]
with $H_j = p_j$ for all sink nodes. Nodes with larger $H_j$ are the \emph{bottlenecks} of the execution batch: delaying them propagates idle time through the entire downstream dependency chain. Accordingly, the scheduling priority of each job is set to $\pi_j = H_j$. In the figure, critical-path nodes are visually distinguished by their annotated height values, allowing the scheduler to immediately identify which jobs must be dispatched first to keep all threads maximally occupied.

As a baseline reference, Figure \ref{fig:list_scheduling}(b) depicts execution under a \emph{single-threaded FIFO} discipline, where orders are processed one at a time in strict arrival order on a single thread $X_1$. All nine orders are serialized, and the makespan equals the sum of all individual processing times. This baseline illustrates the theoretical lower bound on parallelism: even if more threads are available, without intelligent scheduling, the system degenerates toward this worst case.

Figure \ref{fig:list_scheduling}(d) shows what happens when a \emph{greedy, uncoordinated parallel} strategy is applied: multiple threads ($X_1$, $X_2$, $X_3$) are launched simultaneously, and each idle thread simply picks the first pending order from the queue---but only if all predecessors in the DAG have already completed. This ``blocked-activation'' model respects precedence but does so reactively: threads that cannot acquire the necessary inventory lock are suspended and remain idle, causing \emph{thread starvation}. As visible in the timeline, several threads sit idle while waiting for upstream jobs to release shared SKU locks, inflating $C_{\max}$ well beyond the theoretical optimum. In production environments with hundreds of thousands of coupled orders, this idle-wait accumulation drives makespan to several hours.

Figure \ref{fig:list_scheduling}(f) shows the execution timeline under our List Scheduling algorithm. By dispatching jobs in descending order of $\pi_j$ from a dynamically maintained Ready Queue $Q_t$ (jobs whose all predecessors have completed), the scheduler proactively deals with the idle-wait problem seen in EC.1(d). At every scheduling epoch, any idle thread immediately receives the highest-priority available job; upon a job's completion, its successors are instantly unlocked and inserted into $Q_t$. Comparing EC.1(f) with EC.1(d), the reduction in $C_{\max}$ is clearly visible: the same set of nine orders is completed in fewer time units, with lower idle time.

\vspace{-0.6cm}
\vspace{-1cm}

The complete procedure is summarized in Algorithm~\ref{alg:list-scheduling}.

\begin{algorithm}[htbp]
\footnotesize
\setlength{\baselineskip}{0.9\baselineskip}
\caption{List Scheduling for Coupled Orders}
\label{alg:list-scheduling}

\begin{algorithmic}[1]

\Require
    \State Order set $\mathcal{J} = \{j_1, j_2, \ldots, j_N\}$ with processing costs $\{p_j\}_{j \in \mathcal{J}}$
    \State Precedence relation $\prec$ over $\mathcal{J}$ (induced by shared-SKU precedence constraints)
    \State Machine set $\mathcal{M} = \{m_1, m_2, \ldots, m_k\}$ (available CPU threads)

\Ensure
    \State A feasible schedule assignment for all jobs in $\mathcal{J}$
    \State Minimized makespan $C_{\max} = \min\left(\max_{k \in \mathcal{M}} T_k\right)$

\Procedure{ListScheduling}{}

    \State $\triangleright$ \textbf{Phase 1: DAG Construction}
    \State Initialize directed acyclic graph $G = (V, E)$ with $V \leftarrow \mathcal{J},\ E \leftarrow \emptyset$
    \For{each pair $(i, j) \in \mathcal{J} \times \mathcal{J}$}
        \If{$i \prec j$} \Comment{$i$ must release SKU lock before $j$ can proceed}
            \State $E \leftarrow E \cup \{(i, j)\}$
        \EndIf
    \EndFor

    \State $\triangleright$ \textbf{Phase 2: Critical-Path Height Computation}
    \State Compute topological ordering $\sigma$ of $G$
    \For{each node $j$ in reverse order of $\sigma$}
        \If{$j$ has no successors in $G$}
            \State $H_j \leftarrow p_j$
        \Else
            \State $H_j \leftarrow p_j + \max_{\{k \mid (j,k) \in E\}} H_k$
        \EndIf
        \State $\pi_j \leftarrow H_j$ \Comment{Set scheduling priority to critical-path height}
    \EndFor

    \State $\triangleright$ \textbf{Phase 3: Ready Queue Initialization}
    \State $Q \leftarrow \{j \in V \mid \deg_{\mathrm{in}}(j) = 0\}$ \Comment{All source nodes with no predecessors}
    \State $\text{AvailableAt}[k] \leftarrow 0$ for all $k \in \mathcal{M}$

    \State $\triangleright$ \textbf{Phase 4: Priority-Based Dispatch Loop}
    \While{$Q \neq \emptyset$ \textbf{or} any $k \in \mathcal{M}$ is still processing}
        \If{$Q \neq \emptyset$ \textbf{and} $\exists$ idle machine $k \in \mathcal{M}$}
            \State $j^* \leftarrow \operatorname{argmax}_{j \in Q}\ \pi_j$ \Comment{Pull highest-priority ready job}
            \State Assign $j^*$ to machine $k$;\ remove $j^*$ from $Q$
        \EndIf
        \State \textit{// Upon completion of $j^*$: unlock successors}
        \For{each successor $s$ of $j^*$ in $G$}
            \State $\deg_{\mathrm{in}}(s) \leftarrow \deg_{\mathrm{in}}(s) - 1$
            \If{$\deg_{\mathrm{in}}(s) = 0$}
                \State $Q \leftarrow Q \cup \{s\}$ \Comment{Successor unblocked; enqueue immediately}
            \EndIf
        \EndFor
        \State Remove $j^*$ from $G$
    \EndWhile

\EndProcedure
\end{algorithmic}
\end{algorithm}

\section{A Case to Show Intelligent Diagnosis}
\label{appendix:intel_diagnosis}

In this section, We illustrate the process of intelligent diagnosis with a real business case from JD.com. In this instance, the core metric of concern for the business department is the warehouse and transportation cost ratio (cost of warehouse and transportation over gross merchandise volume). 

\vspace{-0.5cm}
\begin{figure}[H]
    \centering
    \includegraphics[width=\linewidth]{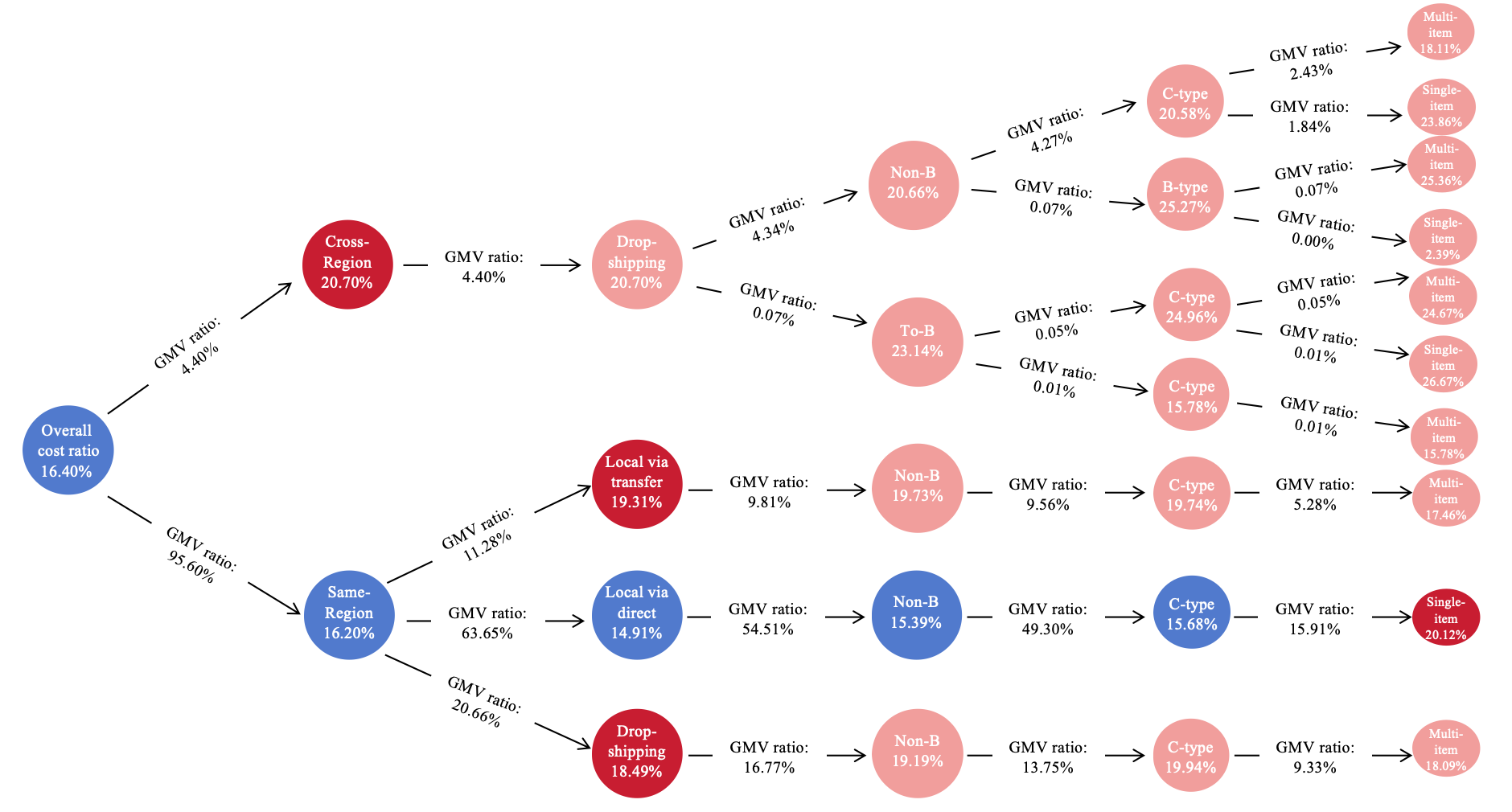}
    \caption{A Case of Intelligent Diagnosis. Blue nodes represent normal states, while light red and dark red nodes both indicate anomalies, with dark red denoting the ultimately identified root cause.}
    \label{fig:int_dia_case}
\end{figure}
\vspace{-1cm}

The diagnostic analysis proceeds along a predefined hierarchical factor sequence:
Cross-Region Type $\rightarrow$ Fulfillment Type $\rightarrow$ Customer Type $\rightarrow$ Billing Type $\rightarrow$ Order Structure. These factors are defined as follows: (1) Cross-Region Type: Distinguished shipments based on the relationship between the destination Regional Distribution Center (RDC) and the shipping DC's parent RDC (cross-region or not). (2) Fulfillment Type: Direct local fulfillment, local fulfillment via transfer, and drop-shipping. (3) Customer Type: Categorized as either enterprise or other. (4) Billing Type: C-type or B-type. (5) Order Structure: Classified as single-item versus multi-item orders. (6) The anomaly criteria is twofold (in this example): 1) any cross-regional shipment is automatically flagged as an anomaly, and 2) any node is flagged if its average cost exceeds the overall mean and its associated business volume (measured by Gross Merchandise Volume, GMV) accounts for more than 5\% of the total. Based on this framework, historical order fulfillment data for the business department is extracted. The system then calculates the average cost ratio at each level of the factor hierarchy and identifies anomalous nodes, as shown in Figure \ref{fig:int_dia_case}. The subsequent root cause analysis reveals that elevated costs are primarily associated with cross-regional shipments, local fulfillment via transfer, drop-shipping, and single-item orders. Following a multi-dimensional assessment to quantify the optimization potential for each of these drivers (summarized in Table \ref{tab:Intelligent_Dia}), the findings and strategy insights are fed back into the simulation to inform the optimization of next planning period. 

\begin{center}
\begin{table}[H]
\setlength{\tabcolsep}{2pt}
\renewcommand{\arraystretch}{0.7}
\caption{Optimization Potential Analysis (TIP means Theoretical Improvement Gap)}
\label{tab:Intelligent_Dia}
\begin{center}
\scalebox{0.8}{
\begin{tabular}{cccccc}
    \toprule
    \textbf{Factors} & \textbf{Status-Quo } & \textbf{TIP} & \textbf{Optimized} & \textbf{Suggestions} & \textbf{Weight}\\
    \toprule
    Cross-Regional & 20.7\% & 27,072,558 & 13.5\% & \makecell[c]{Stocking, cross-regional charges,\\ coordinated warehouse transfer} & \makecell[c]{0.3\%\\ =7.2\% $\times$ GMV} \\
    Drop-Shipping & 18.5\% & 157,931,974 & 9.6\% & Direct delivery, drop-shipping charges & 1.8\% \\ 
    Local via Transfer & 19.3\% & 53,207,648 & 13.8\% & Direct delivery, intra-network transfer charges & 0.6\% \\
    Single-Item Order & 20.1\% & -- & -- & -- & -- \\ 
   \toprule
\end{tabular}
}
\end{center}
\end{table}
\end{center}

\vspace{-.5cm}
In practice, the application of this technique has already yielded an annualized cost reduction of 2.6 million dollars in warehousing and distribution for this single business unit. Extrapolating from this success and benchmarking against H1 2022 revenue figures, a full-scale application across all retail divisions is projected to save an estimated 128.3 million dollars in annual warehousing and distribution costs. Such cost reductions are strategically significant, as they directly enhance the market competitiveness of our product categories and enable the capture of a larger market share.